\documentclass[letterpaper]{article} 
\usepackage[preprint]{aaai2027}  
\nocopyright
\usepackage[hyphens]{url}  
\usepackage{graphicx} 
\usepackage{natbib}  
\usepackage{caption} 
\usepackage{algorithm}
\usepackage{algorithmicx}
\usepackage{xspace}
\usepackage{amsmath}
\usepackage[most,skins,theorems]{tcolorbox}
\usepackage{newfloat}
\usepackage{listings}
\usepackage{pifont}
\usepackage{amssymb}
\usepackage{multirow}
\usepackage[table]{xcolor}
\usepackage{tabularx}
\usepackage{algpseudocode}
\tcbuselibrary{listings, breakable}

\usepackage[most,skins,theorems]{tcolorbox}

\DeclareCaptionStyle{ruled}{labelfont=normalfont,labelsep=colon,strut=off} 
\floatstyle{ruled}
\newfloat{listing}{tb}{lst}{}
\floatname{listing}{Listing}

\usepackage{booktabs}

\newcommand{\benchname}{MemU-Bench\xspace}
\newcommand{\methodname}{AppMem\xspace}

\title{\benchname: Benchmarking GUI Agents on Long-Horizon User Interactions}
\author{
    Yizhou Liu\textsuperscript{\rm 1}, Zhengxi Lu\textsuperscript{\rm 1}, Fei Tang\textsuperscript{\rm 1}, Yiwen Qiu\textsuperscript{\rm 1}, Yong Du\textsuperscript{\rm 1}, Tongbo Chen\textsuperscript{\rm 1}, \\
    Bofan Chen\textsuperscript{\rm 1}, Tao Jiang\textsuperscript{\rm 2}, Wenhao Xu\textsuperscript{\rm 2}, Yunfeng Shu\textsuperscript{\rm 1}, Wenqi Zhang\textsuperscript{\rm 1}, \\
    Weiming Lu\textsuperscript{\rm 1}, Jun Xiao\textsuperscript{\rm 1}, Yueting Zhuang\textsuperscript{\rm 1}, Yongliang Shen\textsuperscript{\rm 1}\textsuperscript{$\dagger$}
}
\affiliations{
    \textsuperscript{\rm 1}Zhejiang University, \textsuperscript{\rm 2}Ant Group
}

\begin{document}

\maketitle

\begin{abstract}
Existing Graphical User Interface (GUI) benchmarks predominantly rely on precise and stateless user instructions. In everyday usage, however, instructions are often \emph{vague and history-dependent}: users omit the exact details of their past operations because they no longer remember them, leaving the agent to recover the missing context from long and noisy interaction histories. To bridge this evaluation gap, we introduce \textbf{\benchname}, a mobile benchmark of \textbf{118 vague tasks} grounded in 29 long-horizon user trajectories, where agents must distill raw interaction streams into memory and retrieve the missing context on demand, with task outcomes verified against serialized application databases. Alongside Success Rate, we incorporate Interaction Efficiency (IE) and Success weighted by Path Length (SPL) to expose agents that succeed by exhaustively searching across different applications rather than by \emph{genuinely recalling} the history. Evaluations show that without user history, state-of-the-art systems struggle under such ambiguity: the best-performing Gemini-3.1-Pro achieves an SPL of only 16.4\%, and end-to-end GUI agents fall below 5\%. To address this, we further propose \textbf{\methodname}, a training-free framework that distills user history into a \emph{structured, app-centric} memory bank, tracks interrupted intents across application switches, and selectively retrieves relevant memories via target-app inference. \methodname raises the SPL of GPT-5.5 from 15.7\% to 66.9\% (\textbf{+51.2\%}), outperforms a naive summary baseline by 9.3\%, and delivers consistent gains across evaluated backbones.
\end{abstract}


\begin{figure}[ht]
    \centering
    \includegraphics[width=0.98\linewidth]{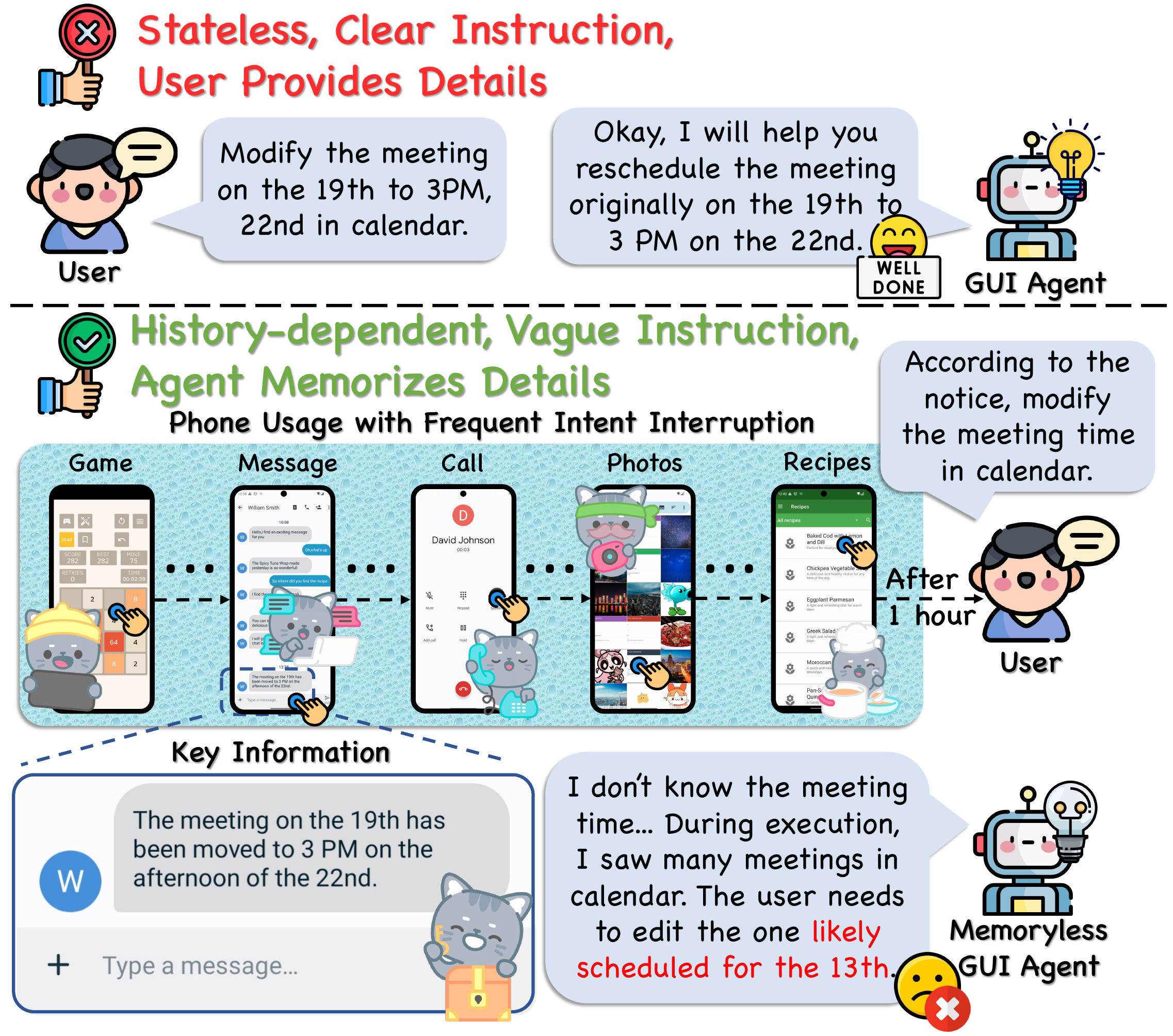}
    \caption{Existing GUI benchmarks (top) evaluate agents on stateless tasks with precise instructions. \benchname (bottom) instead issues vague instructions whose missing context must be recovered from the long user history.
}
    \label{fig:intro}
\end{figure}

\section{Introduction}
\label{sec:intro}

\newcommand{\greycheck}{\textcolor{black!30}{\ding{51}}}   
\newcommand{\greencheck}{\textcolor{green!60!black}{\ding{51}}} 
\newcommand{\redcross}{\textcolor{red!90!black}{\ding{55}}}     

\begin{table*}[!t]
\centering

\footnotesize
\setlength{\tabcolsep}{5pt} 
\resizebox{0.9\textwidth}{!}{%
\begin{tabular}{lcccccc} 
\toprule
\textbf{Benchmark / Dataset} & 
\textbf{\begin{tabular}[c]{@{}c@{}}Vague \\ Ins.\end{tabular}} & 
\textbf{\begin{tabular}[c]{@{}c@{}}User \\ His.\end{tabular}} & 
\textbf{Online} & 
\textbf{\begin{tabular}[c]{@{}c@{}}Missing Info.\\ Acquisition\end{tabular}} & 
\textbf{\begin{tabular}[c]{@{}c@{}}User His.\\Modality\end{tabular}} & 
\textbf{\begin{tabular}[c]{@{}c@{}}Task Target\\ \end{tabular}} \\
\midrule

AITW \citep{aitw}                      & \redcross   & \redcross    & \redcross   & \redcross  & \redcross & GUI Execution \\
Android Control \citep{androidcontrol} & \redcross   & \redcross    & \redcross   & \redcross  & \redcross & GUI Execution \\
AndroidWorld \citep{androidworld}       & \redcross   & \redcross    & \greencheck & \redcross  & \redcross & GUI Execution \\
AndroidLab \citep{androidlab}           & \redcross   & \redcross    & \greencheck & \redcross  & \redcross & GUI Execution \\
MobileWorld \citep{mobileworld}         & \greencheck  & \redcross    & \greencheck & Asking     & \redcross & GUI Execution \\
PIRA-Bench \citep{pira-bench}           & \redcross   & \greencheck  & \redcross   & \redcross  & I & Intent Recommendation \\
AndroidIntent \citep{personalalign}     & \greencheck & \greencheck    & \redcross   & Preference & T & Intent Alignment \\
MemGUI-Bench \citep{memgui-bench}       & \redcross   & \redcross    & \greencheck & \redcross  & \redcross & GUI Execution \\
\textbf{\benchname (Ours)}                     & \greencheck & \greencheck  & \greencheck & History Details    & T+I & History Memory \& GUI Execution \\

\bottomrule
\end{tabular}
}
\caption{Comparison of different benchmarks and datasets. \greencheck incorporated; \redcross not incorporated. T=Text, I=Image}
\label{tab:benchmark_comparison}
\end{table*}
Graphical User Interface (GUI) agents built on Multimodal Large Language Models~\cite{Qwen2-VL, Qwen2.5-VL, Qwen3-VL} are rapidly evolving into general-purpose digital assistants that translate natural language instructions into executable action sequences on real devices~\cite{gui_agents_survey, osagents, seeclick, showui, ui-tars, ui-r1, ui-s1, gui-g2}.

Existing benchmarks almost exclusively evaluate GUI agents under precise, self-contained instructions~\cite{aitw, androidworld, osworld}. Everyday usage rarely looks like this. Since users seldom remember the exact details of their past interactions, they tend to issue \emph{vague, history-dependent} commands such as ``modify the meeting time'', which cannot be resolved unless the agent knows which meeting was created, when, and in which application (Figure~\ref{fig:intro}). Unlike preference-driven personalization, the missing context here consists of \textbf{objective operational evidence}: facts that were fully observable in past interactions but are now buried in hundreds of steps of routine phone usage. Completing such tasks therefore demands more than a capable executor. The agent must \textbf{continuously distill its user's interaction stream into memory} and \textbf{retrieve the right piece of it on demand}.

Faithfully measuring this ability imposes three requirements on benchmark design. First, real interaction histories are \emph{long and noisy}: the few decisive state changes are scattered among hundreds of visually redundant screenshots, so the benchmark must \textbf{expose agents to raw histories} rather than pre-cleaned summaries. Second, real users \emph{multitask}: a single intent is often interrupted, resumed later, and interleaved with unrelated activities in other applications, so the trajectories must \textbf{preserve such fragmented intents}. Third, task success alone is \emph{not evidence of memory}: an agent can complete a vague task by exhaustively searching through applications at execution time, so the evaluation must \textbf{distinguish genuine recall from brute-force exploration}.

We introduce \benchname, a mobile benchmark that instantiates these requirements (Table~\ref{tab:benchmark_comparison}). It comprises \textbf{118 vague tasks} grounded in \textbf{29 long-horizon user trajectories} that cover more than 30 applications, with 100 to 800 steps per trajectory. Tasks are stratified by history length into 71 Level-1 and 47 Level-2 tasks to assess memory scalability. Each task follows a three-phase protocol: the agent first builds a memory bank from the raw history, then retrieves key information for the vague instruction, and finally executes the task online. Outcomes are \textbf{verified programmatically against serialized application databases}, complemented by an LLM judge for open-ended text. Besides Success Rate (SR), we incorporate Interaction Efficiency (IE) and Success weighted by Path Length (SPL)~\cite{anderson2018evaluation}, which penalize agents that reach the goal by \emph{wandering across applications} instead of recalling the relevant context.

Evaluations on \benchname expose a consistent weakness of current systems. Without access to user history, the best-performing model, Gemini-3.1-Pro, attains an SPL of only 16.4\%, and end-to-end GUI agents remain below 5\%. Their trajectories degenerate into aimless exploration across applications, which inflates interaction cost even when the task eventually succeeds. \textbf{Long-term memory is thus not an optional extension but a prerequisite} for handling vague instructions in realistic settings.

Guided by these observations, we propose \methodname, a training-free framework that equips GUI agents with long-term user memory. During daily usage, \methodname summarizes the interaction stream through a sliding window into \emph{app-centric memory entries} and tracks the status of each intent, so that events interrupted by application switching are merged into coherent threads instead of fragmented records. Upon receiving a vague instruction, it infers the target applications, retrieves only their associated entries, and iterates until the collected information is sufficient for execution. On \benchname, \methodname raises the SPL of GPT-5.5 from 15.7\% to 66.9\% (\textbf{+51.2\%}), outperforming the naive summary baseline by 9.3\%, and generalizes across backbones (e.g., improving the SR of MAI-UI-8B by 16.9\%).

We summarize our main contributions below:
\begin{itemize}
    \item We present \textbf{\benchname}, a mobile benchmark of 118 vague tasks over 29 long-horizon trajectories that evaluates whether GUI agents can recover missing context from \emph{long, noisy, and interleaved} user histories, with database-level verification and efficiency-aware metrics.

    \item We systematically evaluate closed-source models, open-source models, and end-to-end GUI agents on \benchname, showing that all of them fail under vague instructions without history and that naive summarization helps general models far more than GUI-specialized agents.

    \item We develop \textbf{\methodname}, a training-free memory framework combining \emph{intent-continuation} memory generation with \emph{app-guided retrieval}, improving SPL by up to 51.2\% and delivering consistent gains across evaluated models.
\end{itemize}

\begin{figure*}[!t]
    \centering
    \includegraphics[width=0.90\linewidth]{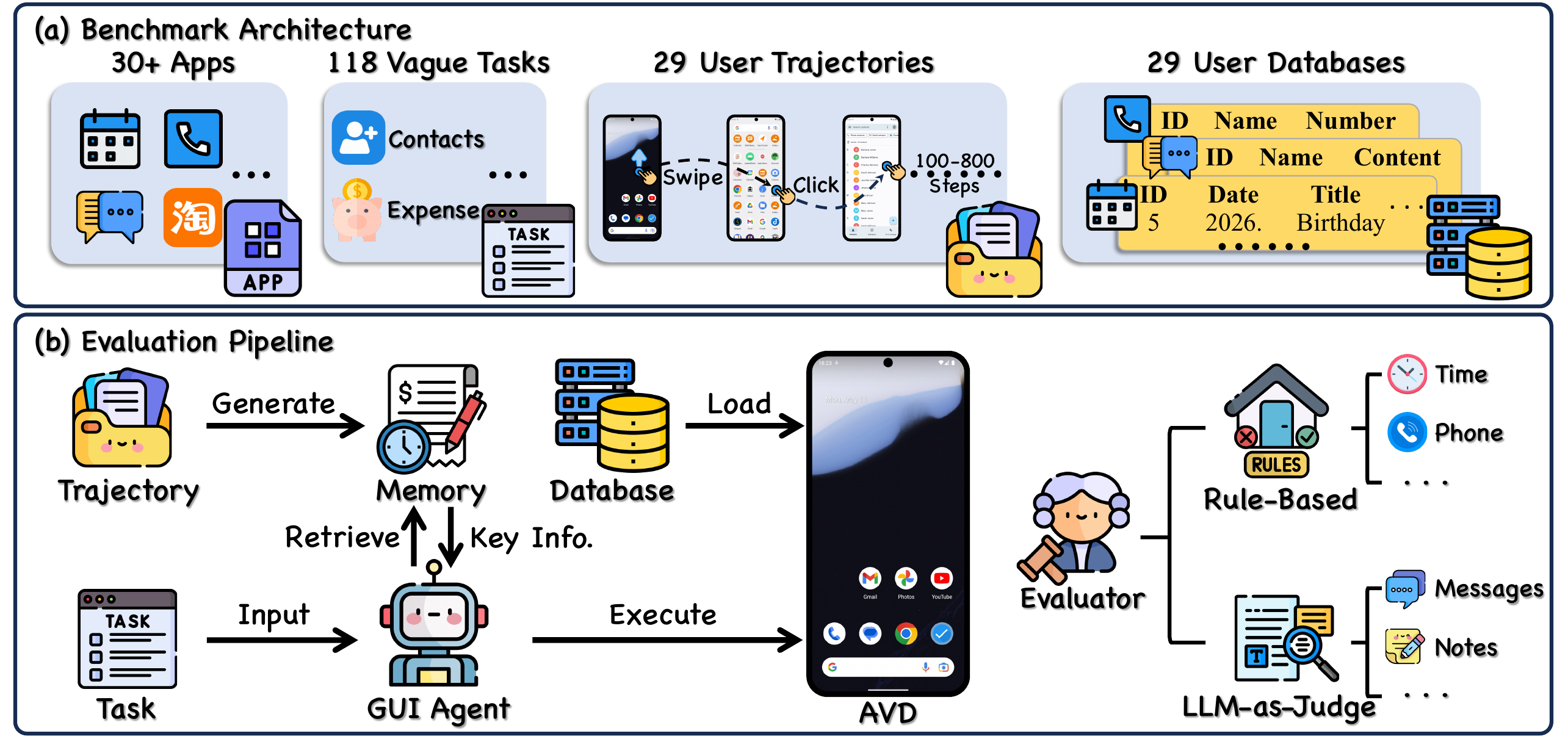}
    \caption{Overview of \benchname. \benchname covers more than 30 applications in user trajectories, encompassing 118 vague tasks, 29 user trajectories and corresponding databases. Prior to execution, GUI agents construct a memory bank from the user history. Upon receiving a vague instruction, it retrieves the key information from its memory and then execute the task.}
    \label{fig:benchmark}
\end{figure*}

\section{Related Work}
\label{sec:related}

\subsection{GUI Agents and Memory Mechanisms}

GUI agents have evolved from script-based automation~\cite{webarena, mobile-agent} to Multimodal Large Language Model agents that execute closed-loop actions directly from pixels~\cite{ui-tars, ui-tars-2}. These agents, however, assume \emph{stateless and precise} instructions~\cite{mai-ui, mobile-agent-v3.5, mobilerl, step-gui, ui-venus, xiaomi-gui-0}: each task is an isolated episode, and prior user activity is discarded. Memory offers a natural remedy and has been studied extensively for LLM agents, which consolidate long dialogue or task histories into retrievable structures~\cite{mem0, memrl, memory_in_agents, memory_in_llm, llm_agent_memory}. These systems, however, operate on \emph{textual} interactions whose semantics are explicit, whereas GUI histories are streams of screenshots and low-level actions with intents fragmented across applications. Existing GUI memory modules sidestep this difficulty by memorizing only transient states within a single task execution~\cite{ui-copilot, se-ga, mem-w, mementogui, stamp}. \methodname instead maintains a persistent, app-centric memory across user's interaction stream.

\subsection{Mobile GUI Benchmarks}
Existing mobile GUI benchmarks follow a task-oriented paradigm: they measure execution success under precise instructions and treat each task as independent of the user's prior history~\cite{aitw, androidcontrol, androidlab, androidworld}. Benchmarks that move beyond this setting relax it only partially. MemGUI-Bench confines memory to a single task~\cite{memgui-bench}, and MobileWorld resolves ambiguity by querying the user in real time rather than by consulting long-term history~\cite{mobileworld}. Personalization benchmarks do exploit history, but only to extract \emph{static preferences} such as habits and tastes~\cite{personalalign, knowu-bench, pspa-bench, pira-bench}. None of them evaluates whether an agent can distill \emph{objective, task-critical facts} from long, noisy, and unstructured daily usage streams. \benchname fills this gap: its vague tasks can only be completed by recovering missing context from raw long-horizon histories, and outcomes are verified against serialized application databases (Table~\ref{tab:benchmark_comparison}).


\section{\benchname}

\subsection{Task Definition}
Each task in \benchname proceeds in three sequential phases, as illustrated in Figure~\ref{fig:benchmark}(b).

\paragraph{Memory Generation.} The GUI agent is initially provided with a historical interaction trajectory $\mathcal{T}$, formulated as:
$$\mathcal{T} = \left( (o^H_1, a^H_1, t_1), (o^H_2, a^H_2, t_2), \dots, (o^H_N, a^H_N, t_N) \right)$$
where $o^H_i$, $a^H_i$, and $t_i$ denote the $i$-th historical screenshot, atomic action (e.g., click coordinates), and absolute interaction time, respectively. Reflecting the intermittent nature of real phone usage, the interval $\Delta t_i = t_{i+1} - t_i$ ranges from under a second to a day.

The GUI agent is then required to condense this long-horizon history into a memory bank $\mathcal{M}$, formalized as:
$$\mathcal{M} = f_{\text{generate}}(\mathcal{T})$$
where $f_{\text{generate}}$ represents the memory mechanism, and $\mathcal{M}$ can manifest in various formats (e.g., natural language summaries, or structured databases).

\paragraph{Key Information Retrieval.} Given a vague instruction $\tilde{I}$, the GUI agent retrieves key information $\kappa$ from $\mathcal{M}$:
$$\kappa = f_{\text{retrieve}}(\tilde{I}, \mathcal{M})$$
where $f_{\text{retrieve}}$ represents the retrieval mechanism.

\paragraph{Online Execution.} Conditioned on $\kappa$ and $\tilde{I}$, the GUI agent executes a sequence of actions within a Markov Decision Process framework. Formally, at each discrete step $t$, it interacts with the mobile environment by sampling an action $a_t$ from its policy $\pi$:

$$a_t \sim \pi(a \mid \tilde{I}, o_t, h_{<t}, \kappa)$$
where $o_t$ denotes the current visual observation (screenshot) at step $t$, and $h_{<t}$ represents the history of executing actions within the current task session.

\subsection{Dataset Construction}

\paragraph{Trajectory Sketches and Task Syntheses.} We first manually design vague tasks together with intent-level trajectory sketches. Each sketch outlines a sequence of high-level user intents for a long-horizon session (e.g., ordering food, then casual gaming) without prescribing specific applications or operation paths. The information required to resolve the final vague task is embedded into the interaction process itself. Operators are therefore free to vary applications and execution paths, which increases trajectory diversity.

\paragraph{Synchronous Trajectory Collection and Granular Logging.} Guided by the sketches, operators interact with an Android Virtual Device emulator to simulate daily phone usage. Rather than following deterministic paths, they introduce realistic noise such as accidental touches and redundant app switching. Sessions are fragmented across multi-hour spans with long inactive gaps, mirroring how people actually use their phones. A background script concurrently captures screenshots and atomic actions at 0.3-second intervals, preserving the complete raw interaction log.

\paragraph{Data Processing and State Serialization.} Consecutive identical screenshots are filtered out, yielding clean, action-interleaved logs. Upon trajectory completion, the emulator state is frozen and serialized: we record the system time and pull the runtime databases (e.g., SQLite files) of all involved applications. Restoring these snapshots before testing reproduces the exact device state left by the operator, so every effect of the historical interaction (a sent message, a modified event) exists in the backend and can be verified against ground-truth records.

\begin{figure}[H]
  \centering
  \begin{minipage}[t]{0.49\linewidth}
    \centering
    \includegraphics[width=\linewidth]{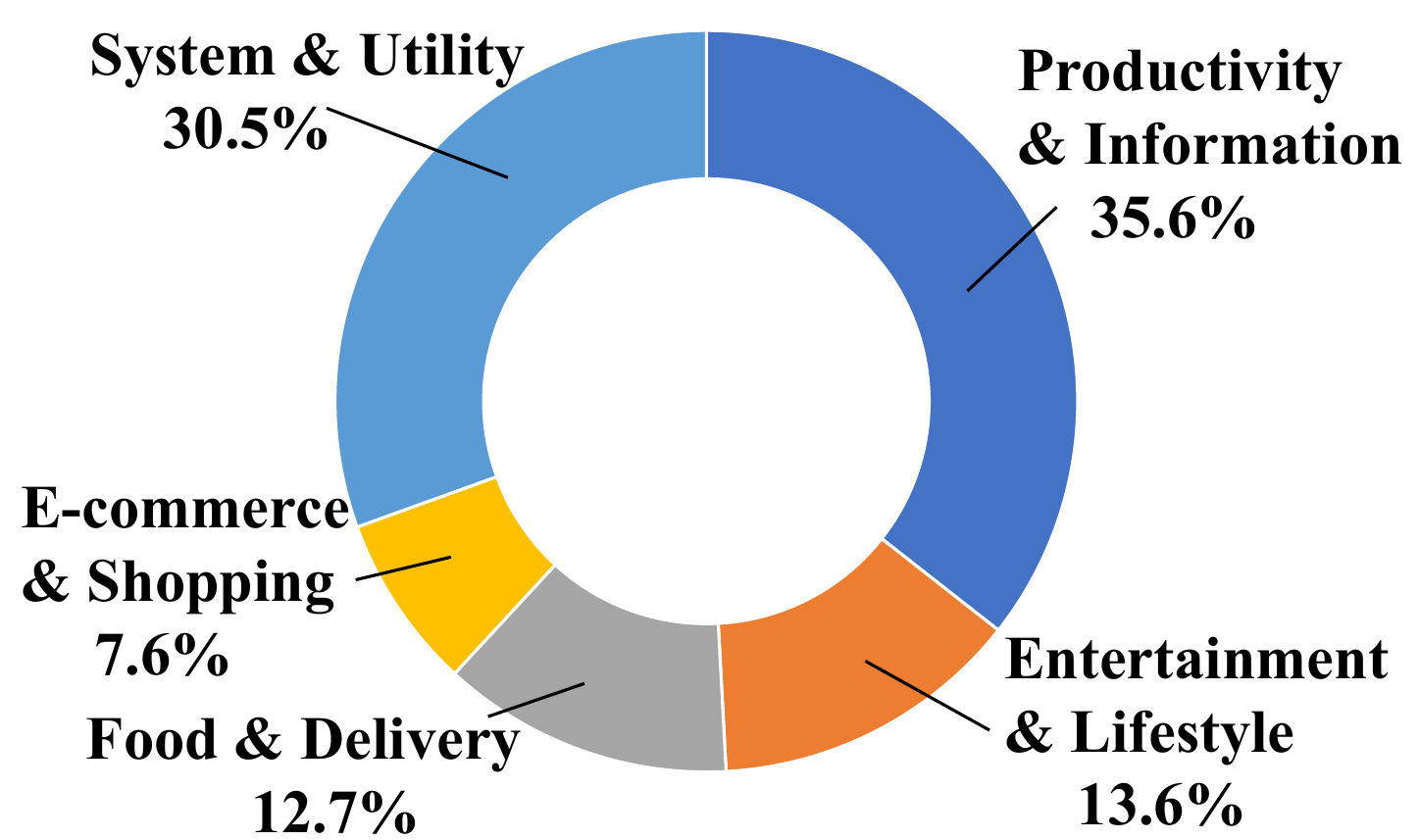}
  \end{minipage}
  \hfill
  \begin{minipage}[t]{0.49\linewidth}
    \centering
    \includegraphics[width=\linewidth]{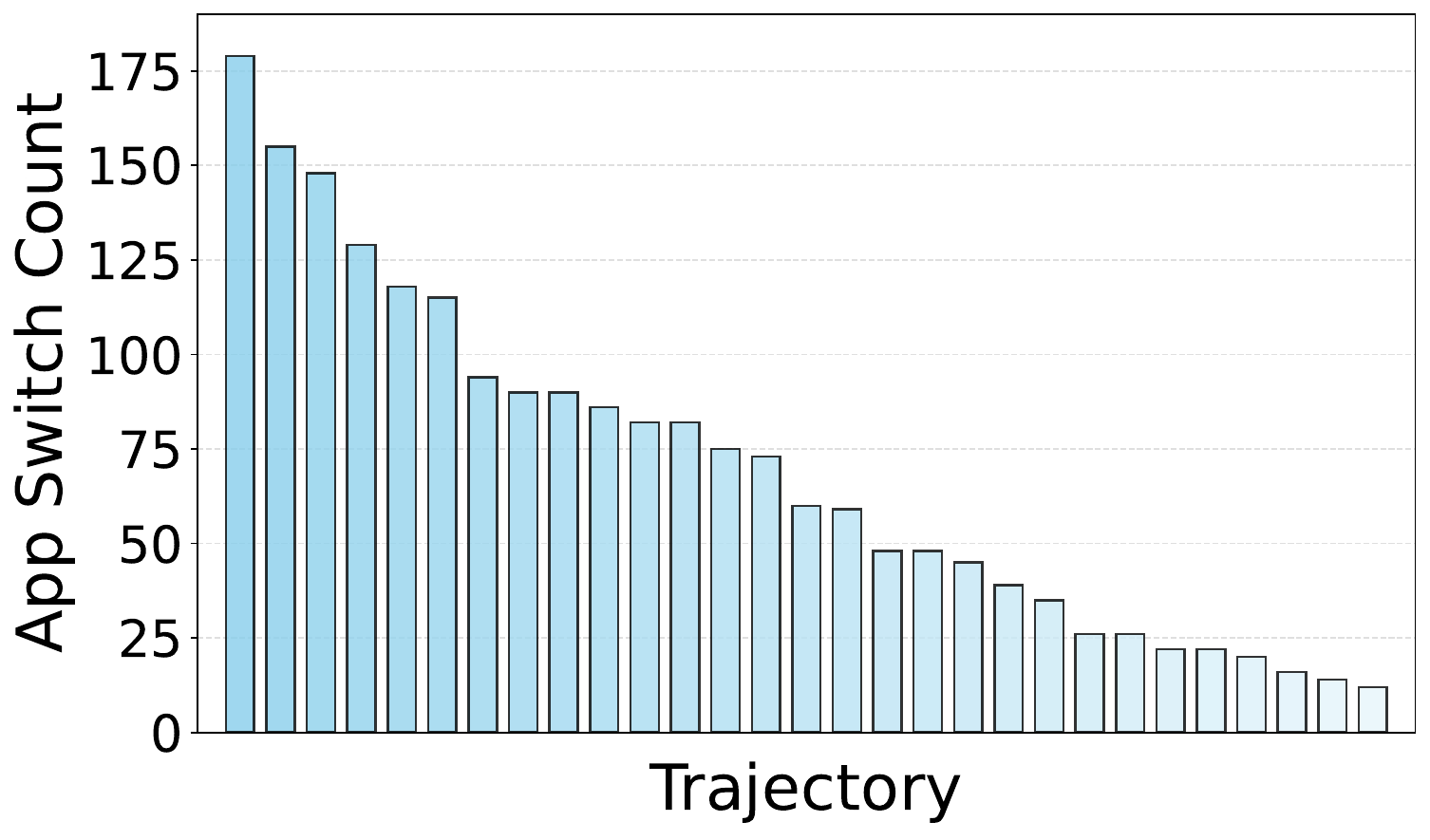}
  \end{minipage}
  \caption{Statistics of \benchname: Left: Task distribution of target applications; Right: Sorted application switch count in user historical trajectories.}
  \label{fig:overall_stats}
\end{figure}

The resulting benchmark contains 29 user trajectories and 118 tasks: 71 Level-1 (L1) tasks with shorter histories (fewer than 400 screenshots) and 47 Level-2 (L2) tasks embedded in longer flows (more than 700 screenshots). As shown in Figure~\ref{fig:overall_stats}, tasks span 5 major application categories, and trajectories cover more than 30 applications, with up to 179 application switches within a single trajectory. Further statistics are provided in the supplementary materials.

\subsection{Evaluation Strategies}
Task success is verified by a cascade of checks, applied in order of decreasing determinism.

\paragraph{Rule-based Verification.} For tasks with deterministic objectives and no open-ended text, we extract the raw backend records from the application's database and verify success conditions programmatically.

\paragraph{LLM-as-Judge.} For tasks involving open-ended text (e.g., messages), deterministic fields such as timestamps are rule-checked, and only the free-text fields are forwarded to an LLM judge for semantic comparison. For a minority of tasks where the database is inaccessible, a VLM evaluates the outcome from screenshots and actions. Restricting the judge to short text fields keeps context small and decisions consistent.

\subsection{Metrics}
\label{metrics}

\paragraph{Success Rate (SR).} SR measures completed task rate:
$$\text{SR} = \frac{|C|}{N}$$
where $N$ represents the total number of tasks, $C$ denotes the set of successfully completed tasks.

\paragraph{Interaction Efficiency (IE).} IE evaluates the agent's operational efficiency, calculated exclusively over successfully completed tasks. IE serves as a proxy to distinguish whether the agent successfully recalls historical context or relies on blind exploration for clues during task execution. Formally,
$$\text{IE} = \frac{1}{|C|} \sum_{i \in C} \frac{S^{\text{golden}}_i}{\max(S^{\text{agent}}_i, S^{\text{golden}}_i)}$$
where $S^{\text{agent}}_i$ and $S^{\text{golden}}_i$ represent the number of steps taken by the GUI agent and human for task $i$, respectively.

\paragraph{Success weighted by Path Length (SPL).} To jointly evaluate task success and interaction efficiency, we adapt Success weighted by Path Length (SPL)~\cite{anderson2018evaluation}:
$$\text{SPL} = \frac{1}{N} \sum_{i=1}^{N} \mathbb{I}[i \in C] \cdot \frac{S^{\text{golden}}_i}{\max(S^{\text{agent}}_i, S^{\text{golden}}_i)}$$
where $\mathbb{I}[i \in C]$ is an indicator function that equals $1$ if task $i$ is successful and $0$ otherwise.


\section{\methodname}
\begin{figure*}[!t]
    \centering
    \includegraphics[width=0.90\linewidth]{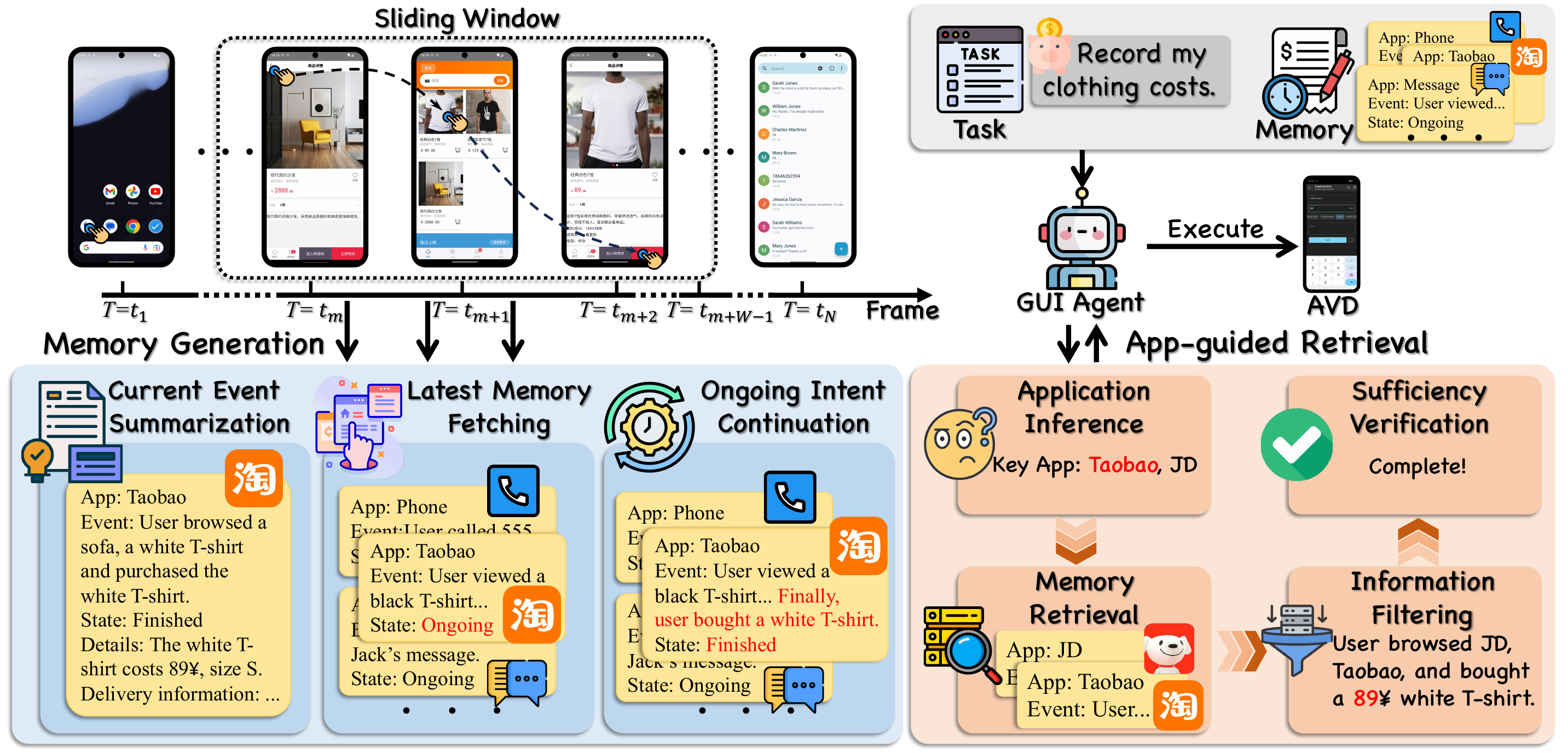}
    \caption{Overview of \methodname. During the memory generation phase, our method concatenates the user’s interrupted historical intents. In the retrieval phase, relevant information is retrieved on an app-centric basis to reduce Token consumption.}
    \label{fig:method}
\end{figure*}

\newcommand{\up}[1]{{\scriptsize (+#1)}}

\newcommand{\down}[1]{}
\newcommand{\redimprove}[1]{\up{#1}}
\newcommand{\improve}[1]{\up{#1}}
\newcommand{\greendecrease}[1]{\down{#1}}
\newcommand{\decrease}[1]{\down{#1}}
\begin{table*}[!t]
\centering

\footnotesize
\begin{tabular}{lccccccc}
\toprule
\multirow{2}{*}{\textbf{Model}} & \multicolumn{2}{c}{\textbf{L1}} & \multicolumn{2}{c}{\textbf{L2}} & \multirow{2}{*}{\textbf{Avg. IE}} & \multirow{2}{*}{\textbf{Avg. SR}} & \multirow{2}{*}{\textbf{Avg. SPL}} \\
\cmidrule(lr){2-3} \cmidrule(lr){4-5}
 & \textbf{SR} & \textbf{SPL} & \textbf{SR} & \textbf{SPL} & & & \\
\midrule

\rowcolor{gray!10} \multicolumn{8}{l}{\textit{Closed-source Models}} \\
Gemini-3.1-Pro~\citep{gemini} & 26.8 & 18.1 & 23.4 & 14.0 & 0.65 & 25.4 & 16.4 \\
\quad w/ Summary & 43.7 & 34.7 & 36.2 & 27.2 & 0.78\improve{0.13} & 40.7\improve{15.3} & 31.7\improve{15.3} \\
\quad w/ \textbf{\methodname (Ours)} & \textbf{64.8} & \textbf{55.2} & \textbf{63.8} & \textbf{53.5} & \textbf{0.85}\redimprove{0.20} & \textbf{64.4}\redimprove{39.0} & \textbf{54.5}\redimprove{38.1} \\
\cmidrule(lr){1-8}
GPT-5.5~\citep{gpt} & 28.1 & 18.1 & 17.0 & 11.9 & 0.66 & 23.7 & 15.7\\
\quad w/ Summary & 73.2 & 63.7 & 57.5 & 48.2 & 0.86\improve{0.20} & 67.0\improve{43.3} & 57.6\improve{41.9} \\
\quad w/ \textbf{\methodname} & \textbf{74.7} & \textbf{68.1} & \textbf{72.3} & \textbf{64.9} & \textbf{0.91}\redimprove{0.25} & \textbf{73.7}\redimprove{50.0} & \textbf{66.9}\redimprove{51.2} \\
\cmidrule(lr){1-8}
Claude-Opus-4.8~\citep{claude} & 16.9 & 14.7 & 19.2 & 13.4 & 0.80 & 17.8 & 14.2 \\
\quad w/ Summary & \textbf{57.8} & \textbf{50.3} & 44.7 & 37.6 & 0.86\improve{0.06} & 52.5\improve{34.7} & 45.3\improve{31.1} \\
\quad w/ \textbf{\methodname} & 56.3 & 49.7 & \textbf{53.2} & \textbf{47.8} & \textbf{0.89}\redimprove{0.09} & \textbf{55.1}\redimprove{37.3} & \textbf{49.0}\redimprove{34.8} \\

\midrule
\rowcolor{gray!10} \multicolumn{8}{l}{\textit{Open-source Models}} \\
Qwen3.5-397B-A17B~\citep{qwen3.5} & 18.3 & 13.3 & 6.4 & 4.3 & 0.72 & 13.6 & 9.7\\
\quad w/ Summary & 40.9 & 35.1 & 27.7 & 20.4 & 0.82\improve{0.10} & 35.6\improve{22.0} & 29.3\improve{19.6} \\
\quad w/ \textbf{\methodname} & \textbf{56.3} & \textbf{50.5} & \textbf{46.8} & \textbf{41.9} & \textbf{0.90}\redimprove{0.18} & \textbf{52.5}\redimprove{38.9} & \textbf{47.1}\redimprove{37.4} \\
\cmidrule(lr){1-8}
Qwen3-VL-32B~\citep{Qwen3-VL} & 2.8 & 1.5 & 2.1 & 0.6 & 0.45 & 2.5 & 1.2 \\
\quad w/ Summary & 9.9 & 7.6 & 8.5 & 5.8 & 0.74\improve{0.29} & 9.3\improve{6.8} & 6.9\improve{5.7} \\
\quad w/ \textbf{\methodname} & \textbf{18.3} & \textbf{15.5} & \textbf{14.9} & \textbf{12.7} & \textbf{0.85}\redimprove{0.40} & \textbf{17.0}\redimprove{14.5} & \textbf{14.4}\redimprove{13.2} \\
\cmidrule(lr){1-8}
Qwen3-VL-8B~\citep{Qwen3-VL} & 4.2 & 3.1 & 4.3 & 2.2 & 0.64 & 4.2 & 2.7 \\
\quad w/ Summary & 15.5 & 11.7 & 2.1 & 1.1 & 0.73\improve{0.09} & 10.2\improve{6.0} & 7.5\improve{4.8} \\
\quad w/ \textbf{\methodname} & \textbf{18.3} & \textbf{13.9} & \textbf{8.5} & \textbf{6.3} & \textbf{0.76}\redimprove{0.12} & \textbf{14.4}\redimprove{10.2}& \textbf{10.9}\redimprove{8.2} \\

\midrule
\rowcolor{gray!10} \multicolumn{8}{l}{\textit{End-to-end GUI Agents}} \\
UI-Venus-1.5-8B~\citep{ui-venus} & 1.4 & 1.4 & 2.1 & 0.7 & 0.67 & 1.7 & 1.1 \\
\quad w/ Summary & 2.8 & 0.9 & 0.0 & 0.0 & 0.30\greendecrease{0.37} & 1.7 & 0.5\greendecrease{0.6} \\
\quad w/ \textbf{\methodname} & \textbf{11.3} & \textbf{8.6} & \textbf{12.8} & \textbf{8.7} & \textbf{0.72}\redimprove{0.05} & \textbf{11.9}\redimprove{10.2} & \textbf{8.6}\redimprove{7.5} \\
\cmidrule(lr){1-8}
MAI-UI-8B~\citep{mai-ui} & 5.6 & 4.2 & 4.3 & 2.2 & 0.67 & 5.1 & 3.4 \\
\quad w/ Summary & 7.0 & 5.3 & 6.4 & 3.2 & 0.65\greendecrease{0.02} & 6.8\improve{1.7} & 4.4\improve{1.0} \\
\quad w/ \textbf{\methodname} & \textbf{21.1} & \textbf{15.8} & \textbf{23.4} & \textbf{16.9} & \textbf{0.74}\redimprove{0.07} & \textbf{22.0}\redimprove{16.9} & \textbf{16.2}\redimprove{12.8} \\
\cmidrule(lr){1-8}
GUI-Owl-1.5-8B~\citep{mobile-agent-v3.5} & 2.8 & 2.5 & 2.1 & 0.5 & \textbf{0.67} & 2.5 & 1.7 \\
\quad w/ Summary & 4.2 & 1.4 & 4.3 & 1.0 & 0.30\greendecrease{0.37} & 4.2\improve{1.7} & 1.3\greendecrease{0.4} \\
\quad w/ \textbf{\methodname} & \textbf{5.6} & \textbf{3.8} & \textbf{10.6} & \textbf{5.0} & 0.56\decrease{0.11} & \textbf{7.6}\redimprove{5.1} & \textbf{4.3}\redimprove{2.6} \\

\bottomrule
\end{tabular}
\caption{Performance comparison on \benchname. SR and SPL are percentages, IE lies in $[0,1]$, and higher is better. L1/L2 denote tasks with shorter/longer histories. Bold marks the best result within each backbone group.}
\label{tab:main}
\end{table*}

\methodname targets the two capabilities: consolidating fragmented intents scattered across applications, and retrieving precisely under vague instructions. As illustrated in Figure~\ref{fig:method}, it slides a fixed-size window over the historical log to build a persistent, \emph{app-centric} memory bank, and answers vague instructions through \emph{app-guided retrieval}.

\subsection{Memory Generation}
To construct the memory bank, \methodname slides a fixed-size window $W$ through the raw historical logs. Let $\mathcal{X} = ((o^H_l, a^H_l), \dots, (o^H_r, a^H_r))$ denote the current window containing consecutive screenshot-action pairs, where $r = \min(l + W - 1, N)$ and $N$ is the total trajectory length. Within each window, a three-stage pipeline transforms the raw traces into structured memory.

\paragraph{Current Event Summarization.} Within each window, \methodname first analyzes screenshots and actions to perform event summarization, yielding a set of memory entries $M = \{m_1, \dots, m_{|M|}\}$ for distinct applications. Each entry $m_j \in M$ is a 6-tuple:
$$m_j = \langle \text{app}_j, e_j, d_j, t_{\text{start}, j}, t_{\text{end}, j}, s_j \rangle$$
where $\text{app}_j$ is the application name, $e_j$ is the event description, $d_j$ denotes on-screen details (e.g., prices), $t_{\text{start}, j}$ and $t_{\text{end}, j}$ are timestamps, and $s_j \in \{\text{ongoing}, \text{finished}\}$ indicates the status of the user's intent.

\paragraph{Latest Memory Fetching.} After summarizing $M$ within the current window, \methodname retrieves historical context to maintain task continuity against application switching. For each new entry $m_j \in M$, it queries the global memory bank $\mathcal{M}$ for the most recent entry of the same application, i.e., $m_i$ with $i = \max\{ k \mid \text{app}_k = \text{app}_j \}$, if such an entry exists.

\paragraph{Ongoing Intent Continuation.} Once historical entries are fetched, \methodname evaluates their operational status to update the global memory bank $\mathcal{M}$. For each new entry $m_j$, if the matching entry $m_i$ is still ongoing, the two entries are merged via $\text{update}(m_i, m_j)$, which synthesizes the past context with the new behaviors:
$$\mathcal{M} \leftarrow \mathcal{M} \setminus \{m_i\} \cup \{\text{update}(m_i, m_j)\}, \quad \text{if } s_i = \text{ongoing}$$
If the historical event is finished or no matching entry exists, $m_j$ is appended as a new entry:$$\mathcal{M} \leftarrow \mathcal{M} \cup \{m_j\}, \quad \text{if } s_i = \text{finished} \lor m_i = \emptyset$$Thus, each application's ongoing activity is kept in a single coherent entry, and completed events are never overwritten.

\subsection{App-guided Retrieval}
Upon receiving a vague instruction $\tilde{I}$, the agent retrieves the missing context from the memory bank $\mathcal{M}$ in four steps.

\paragraph{Application Inference.} Let $\mathcal{A}$ denote the set of distinct applications appearing in $\mathcal{M}$. Based on $\tilde{I}$, the agent infers a subset of relevant applications $\mathcal{A}^* \subseteq \mathcal{A}$.

\paragraph{Memory Retrieval.} Once the target applications $\mathcal{A}^*$ are identified, \methodname filters the global memory bank to form an app-specific sub-bank $\mathcal{M}^*$, narrowing the search space:
$$\mathcal{M}^* = \{m_i \in \mathcal{M} \mid \text{app}_i \in \mathcal{A}^*\}$$

\paragraph{Information Filtering.} Given $\mathcal{M}^*$ and $\tilde{I}$, the agent distills the retrieved entries into the key information $\kappa$:
$$\kappa = f_{\text{filter}}(\mathcal{M}^*, \tilde{I})$$
where $f_{\text{filter}}$ denotes the synthesis function executed by the GUI agent to extract the task-relevant information.

\paragraph{Sufficiency Verification.} Finally, the agent assesses whether $\kappa$ suffices to execute $\tilde{I}$. If so, retrieval terminates, execution begins. Otherwise, the explored applications are removed from the candidate set ($\mathcal{A} \leftarrow \mathcal{A} \setminus \mathcal{A}^*$), and the process returns to Application Inference, for up to $K$ iterations.

\section{Experiments}

\begin{figure*}[t]
  \centering
  \begin{minipage}[b]{0.33\textwidth}
    \centering
    \includegraphics[height=3.5cm]{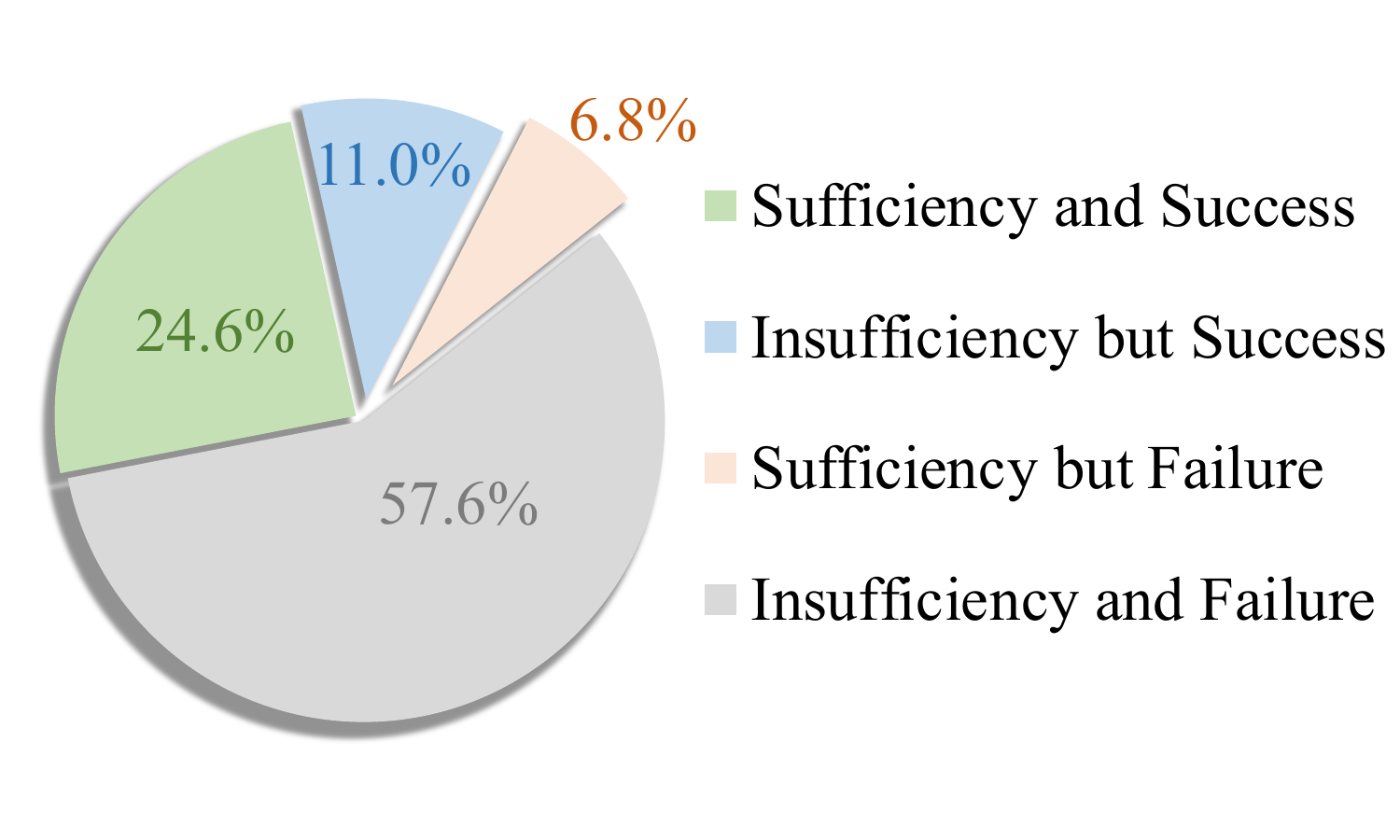}\\[3pt]
    {\footnotesize (a) Error analysis of Summary}
  \end{minipage}
  \hfill
  \begin{minipage}[b]{0.33\textwidth}
    \centering
    \includegraphics[height=3.6cm]{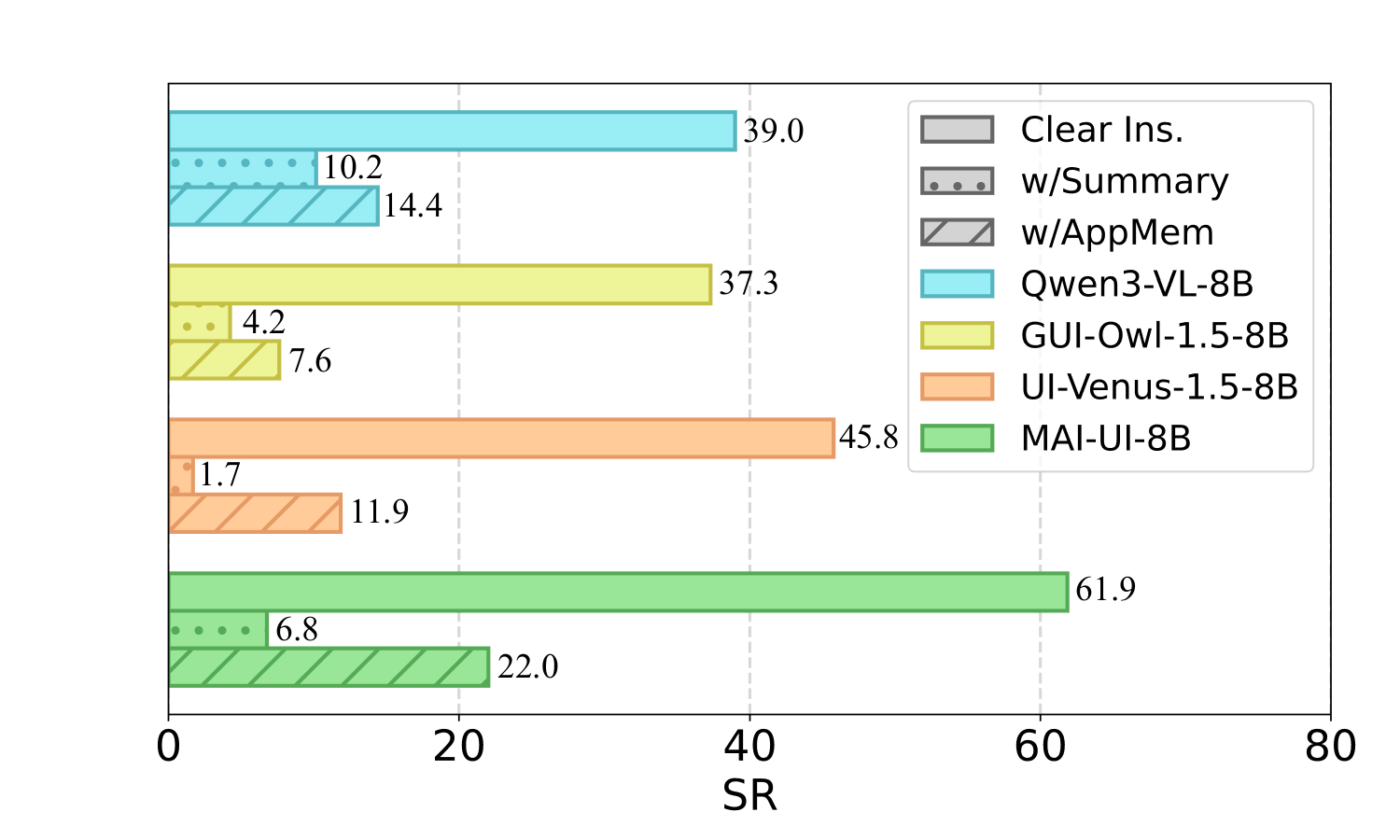}\\[3pt]
    {\footnotesize (b) Precise vs.\ vague instructions}
  \end{minipage}
  \hfill
  \begin{minipage}[b]{0.33\textwidth}
    \centering
    \includegraphics[height=3.6cm]{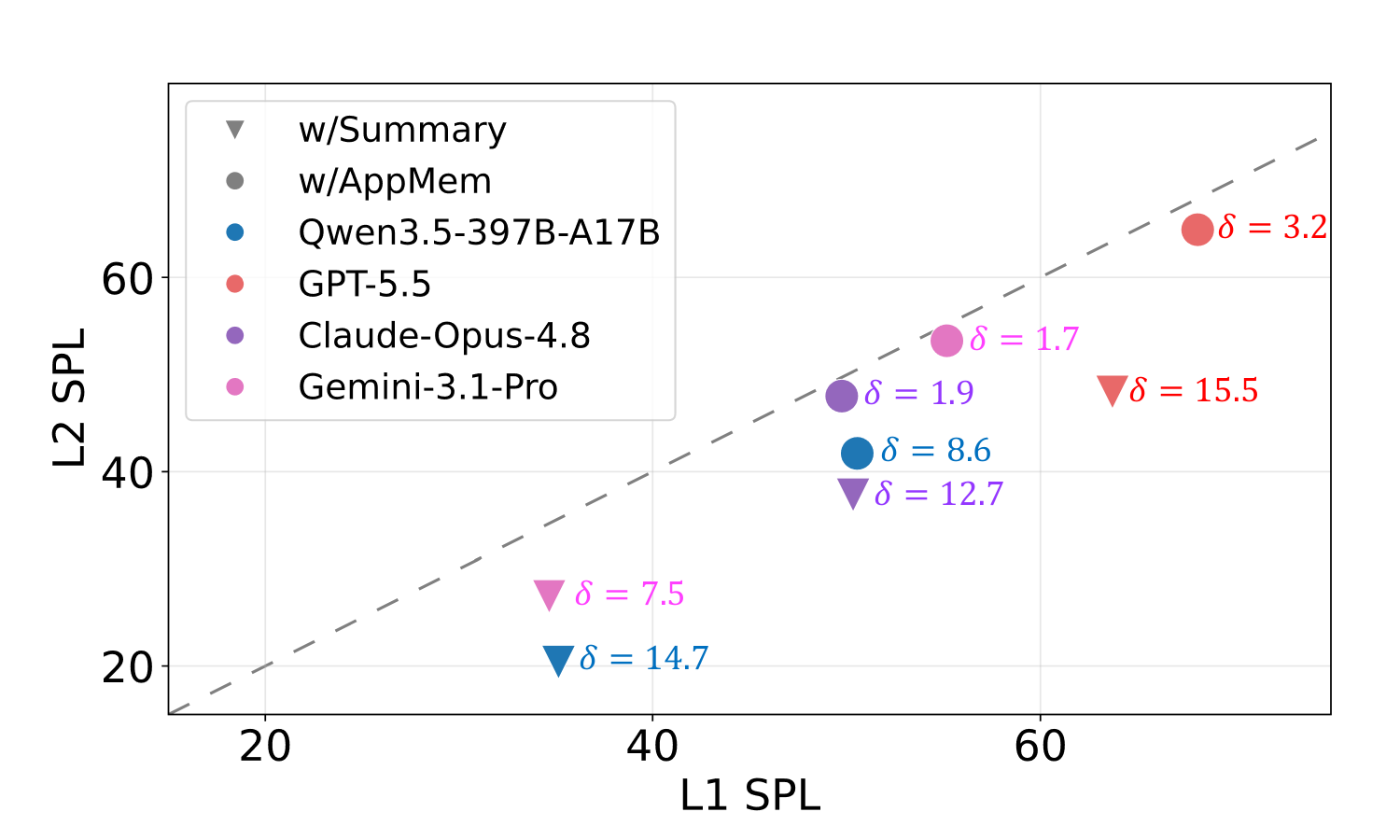}\\[3pt]
    {\footnotesize (c) Robustness to history length}
  \end{minipage}
  \caption{Analysis of the Summary baseline and \methodname: (a) error breakdown by information sufficiency and execution success; (b) end-to-end agents under precise vs.\ vague instructions; (c) L1--L2 SPL gap across models.}
  \label{fig:analysis}
\end{figure*}

\subsection{Experimental Setup}

We evaluate three categories of models on \benchname: (1) \textit{closed-source models}~\cite{gpt, claude, gemini}; (2) \textit{open-source models}~\cite{qwen3.5, Qwen3-VL}; and (3) \textit{end-to-end GUI agents}~\cite{mobile-agent-v3.5, ui-venus, mai-ui}. Each model is tested under two baseline settings: \textit{Vanilla}, which receives no user history, and \textit{Summary}, which applies the same sliding window $W$ to iteratively summarize user behaviors and retrieves key information directly from the summary. All memory and summary operations are performed by the evaluated model itself, without assistance from external models. We set the sliding window size $W = 10$ and the maximum retrieval iterations $K = 3$. Given the cost of processing long-horizon trajectories, each model is evaluated in a single complete run with temperature 0. More implementation details are provided in the supplementary materials.

\subsection{Main Results}

\paragraph{Necessity of User History Memorization.} Without user history, evaluated system fails under vague instructions (Table~\ref{tab:main}). GPT-5.5 attains an average SPL of only 15.7\%, Qwen3.5-397B-A17B reaches 9.7\%, and the strongest end-to-end agent falls to 3.4\%. Vague instructions therefore cannot be treated as isolated inputs: resolving them requires long-term memory of the user's history.

\paragraph{Effectiveness and Generalizability of \methodname.} \methodname improves evaluated backbones, whereas Summary only helps models with strong long-context ability (Table~\ref{tab:main}). With Summary, general models gain substantially (Qwen3.5-397B-A17B: 35.6\% SR), but end-to-end GUI agents barely move (+1.7\% SR): their GUI-specific tuning appears to compromise long-text processing as the history accumulates. \methodname sidesteps this bottleneck by structuring the history into a compact memory bank, raising Qwen3.5-397B-A17B to 52.5\% SR and MAI-UI-8B by +16.9\% SR.

\paragraph{Efficiency.} The efficiency metric reveals an even sharper contrast. Vanilla agents that do succeed still wander (e.g., IE of 0.65 for Gemini-3.1-Pro), and naive summarization can further lower efficiency on weaker backbones, as their limited long-context capability hinders extracting and retaining key information from lengthy historical trajectories (e.g., GUI-Owl-1.5-8B's IE drops from 0.67 to 0.30). \methodname instead improves IE across models, up to 0.91 for GPT-5.5, showing that its compact, targeted memory lets agents act with near-human step economy rather than by brute-force exploration.

\begin{table}[t]
\centering
\footnotesize
\setlength{\tabcolsep}{5pt}
\begin{tabular}{lcccc}
\toprule
 & \textbf{SPL} & \textbf{MS}$\downarrow$ & \textbf{TE}$\downarrow$ & \textbf{CT}$\downarrow$\\
\midrule
\textbf{Full \methodname} & \underline{47.1} & \textbf{8.8K} & \underline{6.1K} & \underline{0.10K} \\
\midrule
\quad w/o OIC & 42.4 & 15.3K & 10.7K & 0.11K \\
\quad w/o AI & \textbf{48.5} & \textbf{8.8K} & 17.8K & 0.15K \\
\quad w/o IF & 41.8 & \textbf{8.8K} & 9.7K & 2.89K \\
\quad w/o SV & 45.3 & \textbf{8.8K} & \textbf{4.9K} & \textbf{0.09K} \\
\bottomrule
\end{tabular}
\caption{Component ablation of \methodname on Qwen3.5-397B-A17B, removing one component at a time. MS: final memory size; TE: retrieval token overhead over successful tasks; CT: context tokens of retrieved information appended at execution. Lower is better for MS, TE, and CT.}
\label{tab:ablation}
\end{table}

\subsection{Ablation Study}
We ablate the components of \methodname on Qwen3.5-397B-A17B (Table~\ref{tab:ablation}). Removing Ongoing Intent Continuation (OIC) lowers SPL by 4.7\% and nearly doubles the memory size (from 8.8K to 15.3K tokens), confirming that stitching interrupted intents produces memory that is both more complete and more compact. For retrieval, without Information Filtering (IF), raw app entries inflate the agent context from 0.10K to 2.89K tokens and SPL drops to 41.8\%. Application Inference (AI) mainly governs cost, cutting retrieval overhead by 66\% (from 17.8K to 6.1K tokens) at nearly the same SPL, while Sufficiency Verification (SV) recovers context scattered across applications at only minor overhead.

\paragraph{RAG.} We replace app-guided retrieval with standard RAG, a bge-m3 retriever over the top-3 memory entries (Table~\ref{tab:rag}). RAG consistently underperforms: on Qwen3.5-397B-A17B it lowers SR and SPL by 9.3\% and 9.6\%, as dense similarity captures neither the heavy lexical overlap nor the fine-grained intent tracking that tasks demand.

\begin{table}[H]
\centering
\footnotesize
\setlength{\tabcolsep}{5pt}
\begin{tabular}{llccc}
\toprule
\textbf{Model} & \textbf{Retrieval} & \textbf{IE} & \textbf{SR} & \textbf{SPL}\\
\midrule
\multirow{2}{*}{GPT-5.5} & RAG & 0.89 & 56.8 & 50.5\\
 & App-guided & \textbf{0.91} & \textbf{73.7} & \textbf{66.9}\\
\midrule
\multirow{2}{*}{Qwen3.5-397B-A17B} & RAG & 0.87 & 43.2 & 37.5\\
 & App-guided & \textbf{0.90} & \textbf{52.5} & \textbf{47.1}\\
\bottomrule
\end{tabular}
\caption{App-guided retrieval vs. standard dense RAG within \methodname, obtained by replacing our retrieval with RAG.}
\label{tab:rag}
\end{table}

\subsection{In-depth Analysis}


\paragraph{Error analysis.} Retrieval, not navigation, is the dominant bottleneck. We manually classify the Summary baseline's outcomes along two axes: whether the retrieved information contains the missing context, and whether online execution succeeds (Figure~\ref{fig:analysis}a). Most tasks (57.6\%) fail due to insufficient information, while only 6.8\% fail despite having it. The remaining cases split into 24.6\% that succeed with sufficient information and 11.0\% that succeed despite insufficient information by exploiting on-screen clues at execution time. We provide a failure case in the supplementary materials.

\paragraph{Performance of end-to-end GUI Agents under clear instructions.} End-to-end agents execute well but cannot use history: their failure is a memory problem, not an execution problem. We additionally evaluate them under explicit instructions that require no history processing (Figure~\ref{fig:analysis}b). The same agents that fail under vague instructions perform far better here, with a gap of 55.1\% for MAI-UI-8B. Their GUI training provides solid execution ability; what they lack is the capacity to summarize and exploit long user histories.

\paragraph{Robustness to history length.} \methodname degrades far less than summarization as histories grow longer (Figure~\ref{fig:analysis}c). Measuring the SPL gap between L1 and L2 tasks ($\delta=\text{SPL}_{\text{L1}}-\text{SPL}_{\text{L2}}$; smaller is more robust), the summary baseline drops sharply (e.g., $\delta=15.5$ for GPT-5.5), whereas \methodname stays stable ($\delta=3.2$), as its structured memory filters noise and sidesteps the long-context bottleneck.

\paragraph{Token efficiency.} \methodname is also cheaper at retrieval time (Table~\ref{tab:retrieval_token}). Through app-guided retrieval, it cuts the retrieval token overhead per successful task by 50.6\% (Gemini-3.1-Pro) relative to the summary baseline.

\begin{table}[H]
  \centering
  \footnotesize
  \begin{tabular}{llc} 
    \toprule
    \textbf{Model} & \textbf{Method} & \textbf{TE} $\downarrow$ \\
    \midrule
    \multirow{2}{*}{Gemini-3.1-Pro} & w/ Summary & 8.5K \\
                                    & w/ \textbf{\methodname} & \textbf{4.2K}($\downarrow$50.6\%) \\
    \cmidrule(lr){1-3}
    \multirow{2}{*}{Claude-Opus-4.8} & w/ Summary & 12.0K \\
                                     & w/ \textbf{\methodname} & \textbf{9.9K}($\downarrow$17.5\%) \\
    \bottomrule
  \end{tabular}
  \caption{Retrieval token efficiency comparison. TE: retrieval token overhead over successful tasks.}
  \label{tab:retrieval_token}
\end{table}

\section{Conclusion}
We introduced \benchname, a mobile GUI benchmark of vague, history-dependent instructions on which state-of-the-art GUI agents fail without user history. Our training-free framework \methodname consolidates histories into app-centric memory entries and retrieves them through app-guided inference, improving both success rate and efficiency across evaluated backbones. More broadly, our results position memory, not execution ability, as the primary bottleneck for long-horizon GUI agents and a key direction for future work.

\appendix


\section{User Historical Trajectories}
\begin{figure}[H]
    \centering
    \includegraphics[width=0.98\linewidth]{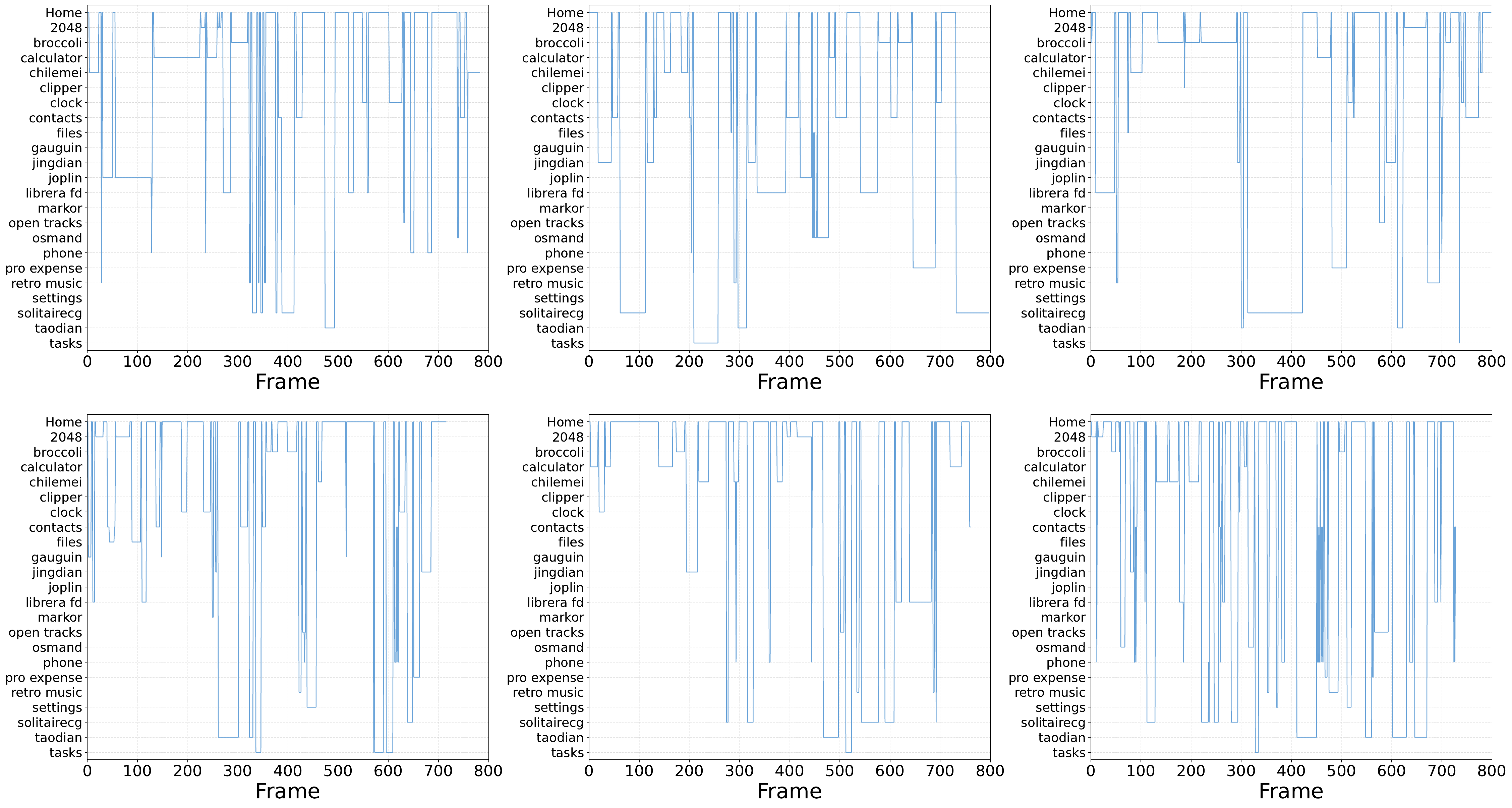}
    \caption{Examples of app transitions across user historical interaction trajectory.}
    \label{fig:app_transtions}
\end{figure}

Figure~\ref{fig:app_transtions} visualizes partial user trajectories to illustrate cross-application switching behaviors over time. The results reveal frequent cross-application switches within the collected user interaction trajectories, which aligns with real-world mobile usage patterns where users frequently shift their intentions during device operation. In addition, significant discrepancies exist across individual users in terms of application usage duration and switching frequency, demonstrating the diversity of \benchname.

\section{Task Distribution}
\label{task_distribution}
Table~\ref{tab:task_distribution} illustrates the task distribution of \benchname. We define the target application as the platform where GUI agents ought to complete the target task. Taking setting an alarm as an example, the specific time constitutes missing information, yet the entire task must be fulfilled within the Clock application. Multiple sources of missing information are contained in \benchname.

\section{Clue-based Inquiry}
We implement an inquiry-based experiment where the agent actively queries users for coarse-grained directional clues without access to user history. This constraint mimics realistic scenarios where users forget fine-grained task details. As shown in Table~\ref{tab:asking}, active inquiry significantly boosts SR (+25.4\%), outperforming the history-based summary baseline. However, it achieves an efficiency of only 0.62 due to the extra operational steps required for information retrieval. In contrast, \methodname outperforms this asking baseline in both SR (52.5\% vs. 39.0\%) and interaction efficiency (0.90 vs. 0.62), validating the effectiveness of our approach.




\begin{table}[H]
\centering
\begin{tabular}{lccc}
\toprule
\textbf{Setting} & \textbf{IE} & \textbf{SR} & \textbf{SPL} \\
\midrule

Qwen3.5-397B-A17B & 0.72 & 13.6 & 9.7 \\
\quad w/ Inquiry & 0.62 & \underline{39.0} & 24.1 \\
\quad w/ Summary & \underline{0.82} & 35.6 & \underline{29.3} \\
\quad w/ \textbf{\methodname} & \textbf{0.90} & \textbf{52.5} & \textbf{47.1} \\

\bottomrule
\end{tabular}
\caption{Performance of Qwen3.5-397B-A17B.}
\label{tab:asking}
\end{table}

\begin{table*}[!t]
\centering
\footnotesize
\begin{tabularx}{\textwidth}{l c X X}
\toprule
\textbf{Target App} & \textbf{Number of Tasks} & \textbf{Example Task} & \textbf{Missing Information}\\
\midrule
Broccoli & 9 & Please open the recipe I want to cook today in broccoli app. & The user wants to cook quick fried rice.\\
\midrule
Chilemei & 4 & Please order a Classic Bolognese Pasta from Handmade Italian Pasta Workshop on chilemei app for Barbara Jones, phone number 13344556677, deliver to her place. & Deliver to Lijin Land Center, Weigongcun Road, Haidian District, Beijing. \\
\midrule
Clock & 6 & I'm cooking according to the recipe I opened earlier, but I forgot how long it takes. Please set an alarm for that many minutes from now in clock app. & It takes 10 minutes to make a Spicy Tuna Wrap.\\
\midrule
Contacts & 6 & Update the phone number for Jennifer Smith in contacts app. & New phone number is 5553300678.\\
\midrule
Files & 1 & Delete the paper I read today, it's too deep for me, in the files app. & The paper the user can't understand is Attention Is All You Need.\\
\midrule
Jingdian & 3 & The cola I ordered before is not enough, please order two more boxes of that cola on jingdian app. & Coke is canned Coke.\\
\midrule
Joplin & 6 & Create a note in joplin app with title 'Reading' and content about the papers I read today. & The user read the Android World paper and Deep Residual Learning for Image Recognition paper.\\
\midrule
Phone & 5 & Call the repairman my friend told me about via phone app. & The phone number is 5552596770.\\
\midrule
Pro Expense & 11 & Record yesterday's dinner expense in pro expense app, name it 'dinner', category is 'social'. & Dinner cost 50 yuan.\\
\midrule
Retro Music & 4 & Please play the song my friend recommended for me in retro music app. & The song is 'Moments'.\\
\midrule
Simple Calendar & 11 & Modify the group meeting time in simple calendar app. & The meeting on May 7th needs to be rescheduled to 4:00 p.m. on May 6th.\\
\midrule
Simple Gallery & 7 & Rename the last receipt I viewed to 'technology' in simple gallery app. & The receipt name is 'receipt\_005.jpg'.\\
\midrule
Simple SMS Messenger & 22 & Share my recorded shopping list with Mary Brown via simple sms messenger app. & The shopping list is recorded in Joplin.\\
\midrule
Solitairecg & 1 & Open the card game I've been playing a lot today. & The game is solitairecg.\\
\midrule
Taodian & 6 & Buy one Women's Lace Floral Cocktail Dress on taodian app for Zhao Si. Payment password is 123456. & Phone number is 15551234567, and address is Building 1, Yard 56, Chaoyang North Road.\\
\midrule
Tasks & 14 & Add a todo in tasks app for buying ingredients, with notes about the ingredients I need for Braised spare ribs. Due date is today. & Main ingredient: 500g pork ribs, choose lean pork ribs with less bones and more meat for better taste. Seasoning: rock sugar, light soy sauce, dark soy sauce, cooking wine, scallions, ginger, garlic, star anise, cinnamon, fragrant leaves.\\
\midrule
Tuantuanwaimai & 2 & Order the burger I recorded before on tuantuanwaimai app and add the remarks. & Remark is: 'Make the Spicy Chicken Leg Burger mildly spicy; for the Double Beef Burger, add extra cheese and omit lettuce'.\\
\midrule
\textbf{Total} & \textbf{118} & & \\
\bottomrule
\end{tabularx}
\caption{Task distribution across different target applications on \benchname.}
\label{tab:task_distribution}
\end{table*}

\section{Token Cost of Memory Generation}
\begin{figure}[H]
  \centering
  \includegraphics[width=0.90\linewidth]{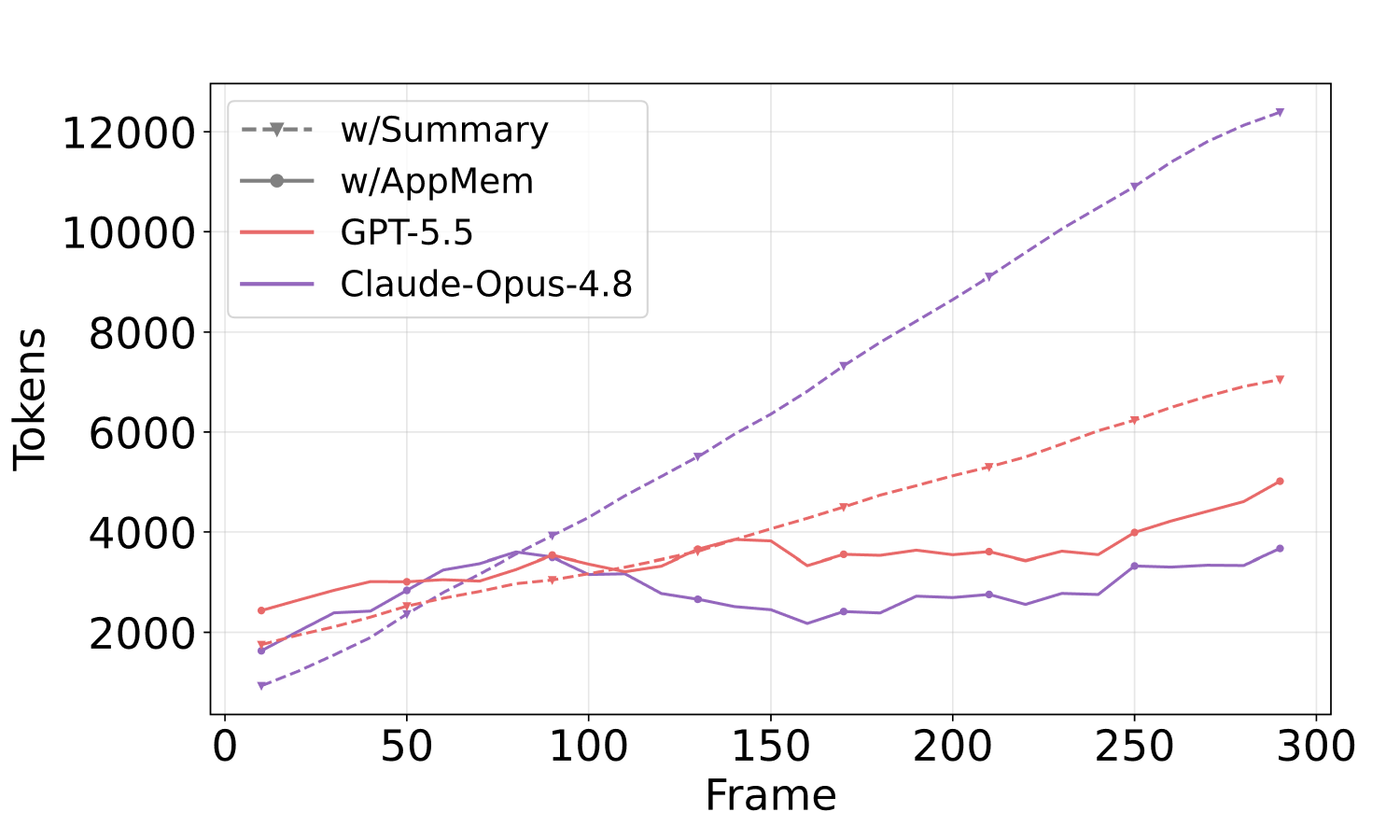} 
  \caption{Token cost during user history processing.}
  \label{fig:generation_token}
\end{figure}

We calculate the per-round text token overhead incurred during user history processing, and the statistical results are presented in Figure~\ref{fig:generation_token}. Summary method generates an updated summary by integrating historical summaries with new user actions at each round, leading to linearly increasing token overhead. In contrast, \methodname constructs a structured memory bank to maintain stable token consumption.

\begin{figure*}[!t]
    \centering
    \includegraphics[width=0.94\linewidth]{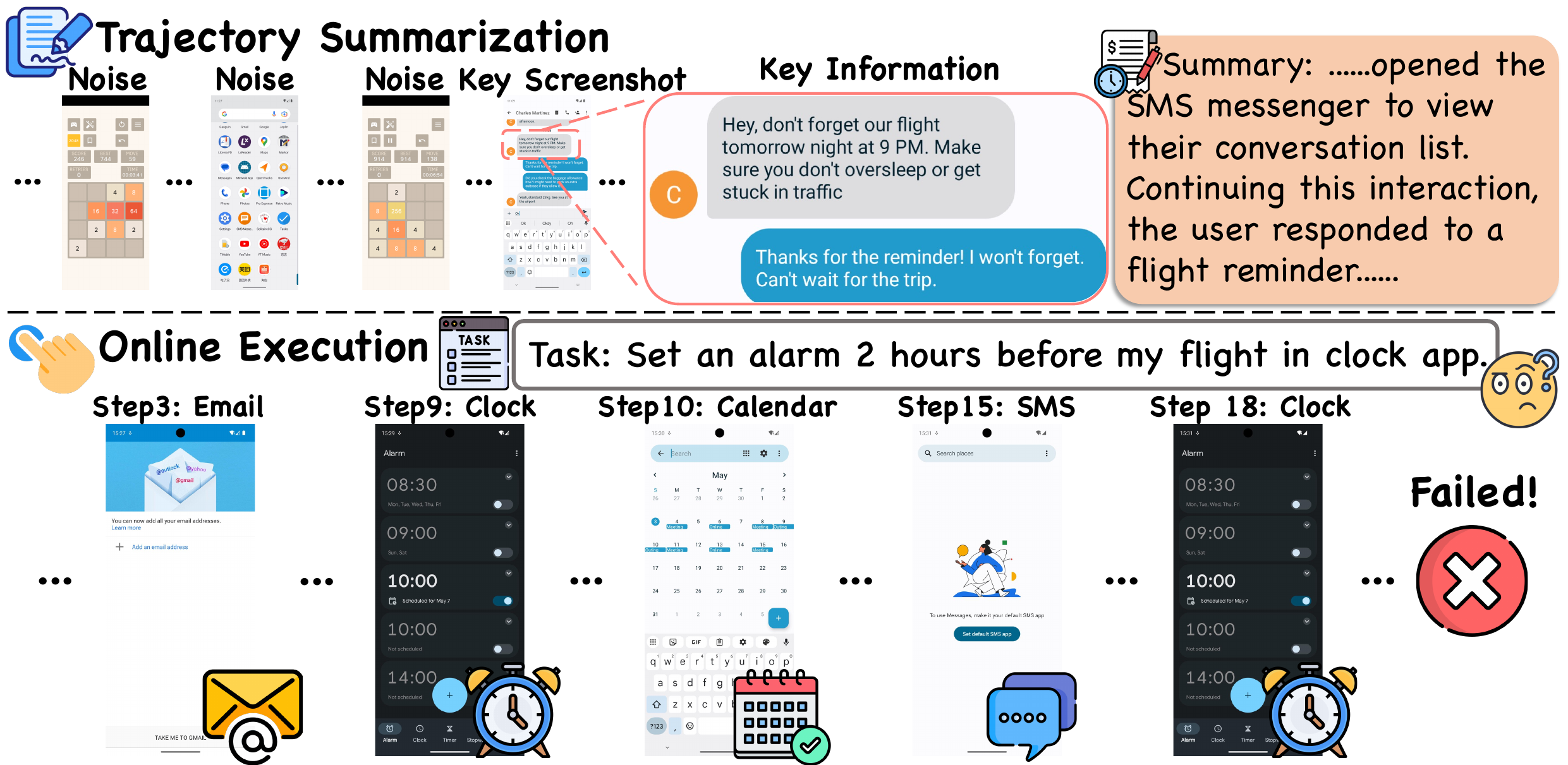}
    \caption{Failure case of the Summary baseline. The summary omits the details required by the vague instruction; the agent then searches multiple applications speculatively and fails.}
    \label{fig:fail_case}
\end{figure*}

\section{Failure Case} 
When the summary omits task-critical details, the agent degenerates into speculative exploration (Figure~\ref{fig:fail_case}). Since the agent cannot know the future instruction while summarizing, lengthy trajectories are compressed lossily; facing the vague instruction, it then searches across applications for the missing content, fails to locate it, and the task fails.

\section{Error Analysis of \methodname}
\begin{figure}[H]
  \centering
  \includegraphics[width=0.90\linewidth]{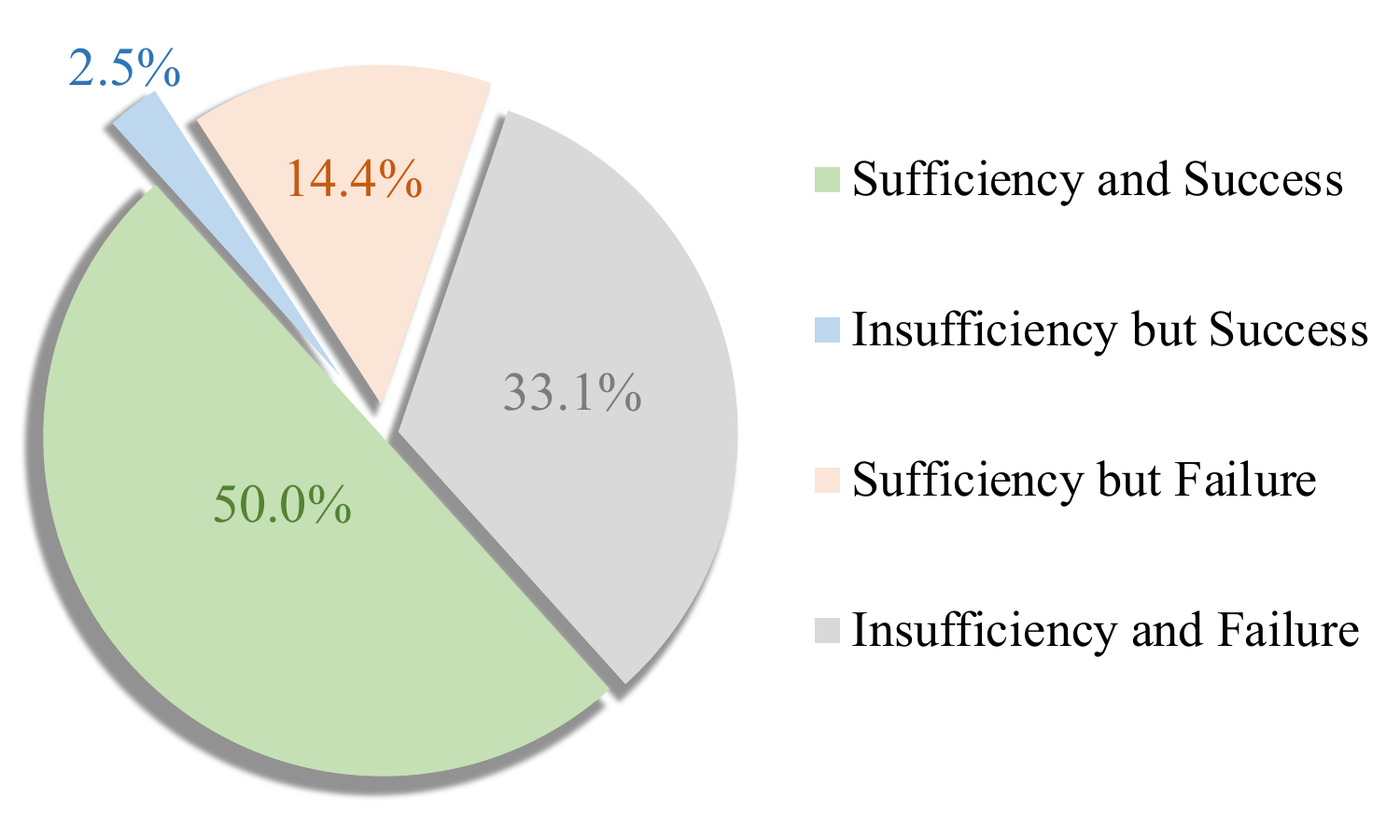}
  \caption{Error analysis of \methodname.}
  \label{fig:error_analysis}
\end{figure}

We manually evaluate Qwen3.5-397B-A17B with \methodname on \benchname (Figure~\ref{fig:error_analysis}) across four categories based on retrieval accuracy and task success. Results show that in 14.4\% tasks, the model retrieved accurate information but failed due to its inherent navigation limitations. Conversely, among the 52.5\% successful tasks, only 2.5\% involved inaccurate retrieval, where the model succeeded by autonomously searching for supplementary data, demonstrating highly effective critical memory retention. Nevertheless, missing information in 33.1\% of tasks highlights remaining limitations.

\begin{figure*}[!t]
    \centering
    \includegraphics[width=0.94\linewidth]{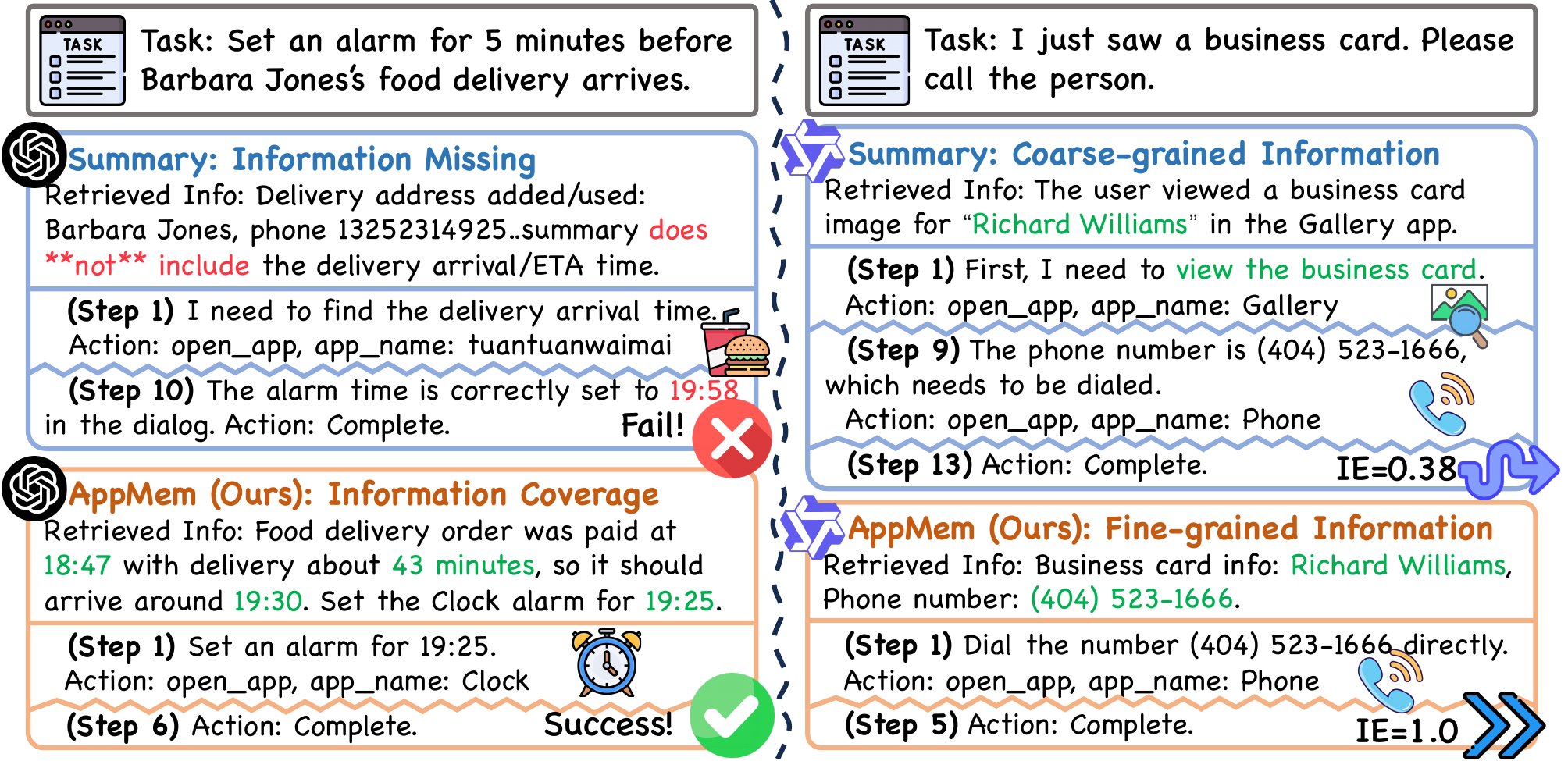}
    \caption{Case study. Left: Information missing and coverage. Right: Coarse-grained and fine-grained information.}
    \label{fig:case_study}
\end{figure*}
\section{Case Analysis of \methodname}

\paragraph{Coverage of Memorized Information.} Figure~\ref{fig:case_study} (left) presents a case where the Summary method fails while our approach succeeds. The summary method fails due to information missing; its summary omits critical details like the exact order and arrival times, forcing the GUI agent into an error-prone search within the app where it retrieves an incorrect time. In contrast, our method meticulously records these temporal details. By directly retrieving the correct schedule from its history without risky on-screen searches, our agent accurately infers the target time and successfully completes the task, demonstrating the vital role of structured memory in resolving real-world ambiguities.

\paragraph{Granularity of Memorized Information.} Figure~\ref{fig:case_study} (right) illustrates the efficiency gap between the two methods. The Summary baseline manages to complete the task but exhibits low efficiency due to its coarse-grained information retention; while it registers the person's name, it fails to record the specific phone number. Consequently, the GUI agent is forced to execute redundant steps to relocate the business card on the phone to get the number before placing the call. In contrast, our method maintains fine-grained information, directly storing the exact phone number in memory. This allows the GUI agent to bypass the time-consuming searching steps and make the call immediately, demonstrating that fine-grained memory retention significantly enhances task execution efficiency.

\section{Evaluation Reliability}
Our evaluation protocol consists of two stages: rule-based verification and LLM-as-Judge. Specifically, 44.9\% of the tasks are evaluated solely through rule-based verification, while the remaining 55.1\% require the LLM-as-Judge to assess open-ended textual outputs. For all LLM-based evaluations, the temperature is set to 0 to ensure deterministic inference and reproducible decisions.

To validate the accuracy of the evaluation protocol, we manually verify the success status of all tasks that require the LLM-as-Judge for Qwen3.5-397B-A17B, and compare the results with those produced by our evaluation protocol. Compared with manual evaluation, our evaluation protocol achieves an accuracy of over 95\%, demonstrating that it provides an accurate and reliable assessment of task success.

\section{Implementation Details}
We set the sliding window size to $W = 10$ (images resized to $672 \times 1492$) and the maximum retrieval rounds $K = 3$. End-to-end GUI agents and Qwen3-VL-8B are evaluated in an end-to-end manner. For other models, We employ the SeeAct-V~\cite{uground} framework, where they act as planners to generate textual actions that are grounded into coordinates by Qwen3-VL-32B-Instruct. Finally, Qwen3.5-35B-A3B serves as the LLM-judge for evaluation. The local evaluation pipeline and agent framework are executed on a workstation equipped with an Apple M4 Pro chip and 24GB of unified memory, running macOS. GUI environments are managed via ADB within Android Studio. Models are invoked through standard OpenAI-compatible API interfaces. The temperature is set to 0 to ensure deterministic inference and reproducibility. To validate the stability of this evaluation protocol, we repeat the evaluation of Qwen3.5-397B-A17B three times under identical settings and observe negligible performance variation across runs (the SR of each run deviates by less than 1.7\% from the mean SR), confirming the low empirical variance of the evaluation protocol. Therefore, a single deterministic run provides a reliable estimate of model performance for benchmarking purposes while substantially reducing API evaluation cost.

\section{\methodname Algorithm}
\label{memory_algorithm}
The implementation of \methodname algorithm is detailed in Algorithm~\ref{alg:memory_generation} and Algorithm~\ref{alg:memory_retrieval}.
\begin{algorithm*}[!t]
\caption{Memory Generation}
\label{alg:memory_generation}
\begin{algorithmic}[1]
\Require User historical trajectory $\mathcal{T} = \left( (o^H_1, a^H_1), (o^H_2, a^H_2), \dots, (o^H_N, a^H_N) \right)$, sliding window size $W$, model $\pi$, prompt for generating events $P_{event}$, prompt for updating memory items $P_{update}$
\Ensure User memory bank $\mathcal{M}$
\For{$l = 1 \mbox{ to } N \mbox{ step } W$}
    \State $r \leftarrow \min(l + W - 1, N)$
    \State $\mathcal{X} \leftarrow \left( (o^H_l, a^H_l), \dots, (o^H_r, a^H_r) \right)$ \Comment{Extract the current window}
    \State $M \leftarrow \pi(\mathcal{X}, P_{event})$ \Comment{Generate the current event}
    \For{\textbf{each} $m \in M$}
        \State $\textit{app} \leftarrow m.\textit{app}$
        \If{$\text{AppExists}(\mathcal{M}, \textit{app})$}
            \State $\tilde{m} \leftarrow \text{FetchLatestMemory}(\mathcal{M}, \textit{app})$ \Comment{Fetch the latest memory item}
            \If{$\tilde{m}.\textit{state} = \text{ongoing}$}
                \State $\mathcal{M} \leftarrow \mathcal{M} \setminus \{ \tilde{m} \} \cup \{ \pi(\tilde{m}, m, P_{update}) \}$ \Comment{Update}
            \Else
                \State $\mathcal{M} \leftarrow \mathcal{M} \cup \{ m \}$ \Comment{Add}
            \EndIf
        \Else
            \State $\mathcal{M} \leftarrow \mathcal{M} \cup \{ m \}$ \Comment{Add}
        \EndIf
    \EndFor
\EndFor
\State \textbf{return} $\mathcal{M}$
\end{algorithmic}
\end{algorithm*}
\begin{algorithm*}[!t]
\caption{App-guided Retrieval}
\label{alg:memory_retrieval}
\begin{algorithmic}[1]
\Require Vague instruction $\tilde{I}$, set of applications $\mathcal{A}$, user memory bank $\mathcal{M}$, model $\pi$, prompt for inferring relevant applications $P_{\textit{infer}}$, prompt for filtering key information $P_{\textit{filter}}$, retrieval rounds $k$
\Ensure Key information $\kappa$
\For{$i = 1 \mbox{ to } k$}
    \If{$i = 1$}
        \State $\mathcal{A}^* \leftarrow \pi(\tilde{I}, \mathcal{A}, P_{\textit{infer}})$ \Comment{Infer applications}
    \Else
        \State $\mathcal{A}^* \leftarrow \pi(\tilde{I}, \mathcal{A}, \epsilon, P_{\textit{infer}})$ \Comment{Infer applications based on the remaining missing information}
    \EndIf
    \State $\mathcal{M}^* \leftarrow \text{Retrieve}(\mathcal{M}, \mathcal{A}^*, \tilde{I})$ \Comment{Fetch relevant memory items}
    \State $\kappa_{\textit{current}} \leftarrow \pi(\mathcal{M}^*, \tilde{I}, \kappa, P_{\textit{filter}})$ \Comment{Filter to obtain key information}
    \State $\kappa \leftarrow \kappa \cup \{ \kappa_{\textit{current}} \}$ \Comment{Accumulate the newly retrieved information}
    \State $\epsilon \leftarrow \text{VerifySufficiency}(\tilde{I}, \kappa)$ \Comment{Identify remaining missing information}
    \If{$\epsilon = \emptyset$}
        \State \textbf{break} \Comment{Stop if no missing information remains}
    \Else
        \State $\mathcal{A} \leftarrow \mathcal{A} \setminus \mathcal{A}^*$ \Comment{Remove inferred applications from the set}
    \EndIf
\EndFor
\State \textbf{return} $\kappa$
\end{algorithmic}
\end{algorithm*}

\section{Prompt Templates}
\label{sec:prompt_template}
We provide prompt templates of \methodname, including Current Event Summarization, Ongoing Intent Continuation, Application Inference, Information Filtering, and Sufficiency Verification.

\begin{tcolorbox}[
    float*=t,                 
    width=\textwidth,         
    title=Prompt Template for Current Event Summarization,
    breakable,
    listing only,
    listing options={
        basicstyle=\ttfamily\small, 
        breaklines=true,            
        columns=flexible
    }
]

You are analyzing a user's operation log on an Android device.\\
I will show you screenshots with the running app, and the actions the user performed before each screenshot. Please analyze the user's activity and generate a 'Memory
Event' for each app.\\
For each app, output in the following format:\\
App: <app name>\\
Event: <A descriptive summary of the user's journey and key interactions. DO NOT use over-generalized summaries like 'User browsed the app'. Preserve the sequence of key
items/categories interacted with (e.g., Browsed T-shirts, compared a yellow one, then switched to sneakers). Capture transitions: If the user changed their mind or compared items, reflect that in the event description.>\\
Details: <Extract all specific data points as a plain text list. Focus on 'Hard Data': 1) Numeric values (verification codes, prices, quantities, durations); 2) Proper nouns (brand names, restaurant names, item titles); 3) Specific attributes (color, size, specifications). If no specific details are found, use None.>\\
State: <finished or ongoing>\\
Start Time: <start time in format 'Apr 22 14:30' or None if unknown>\\
End Time: <end time in format 'Apr 22 14:45' or None if unknown>\\
State Definition:\\
- **finished**: The task within this app reached a terminal point (e.g., payment confirmed, message sent, '=' pressed in calculator, final search result selected).\\
- **ongoing**: The task is incomplete or the user exited prematurely (e.g., form partially filled, still browsing, expression left hanging). If the last operation does
not represent a definitive 'Commit' or 'Result' action, mark as 'ongoing'.\\
\{Example\}\\
Guidelines:\\
1. The current system time is in the upper left corner.\\
2. In the 'Details' field, prioritize recording 'hard information' (numbers, specific names) as this will serve as long-term memory for future retrieval.\\
3. The name of the running app is provided along with the screenshot. Please do not modify the app name on your own. If no app is identified, use None as the app name.\\
Screenshot 1 (app: \{app\_name1\}):\\
\{Screenshot1\}\\
Actions performed:\\
\{Actions1\}\\
Screenshot 2 (app: \{app\_name2\}):\\
\{Screenshot2\}\\
...
\end{tcolorbox}

\begin{tcolorbox}[
    float*=t,                 
    width=\textwidth,         
    title=Prompt Template for Ongoing Intent Continuation,
    breakable,
    listing options={
        basicstyle=\ttfamily\small, 
        breaklines=true,            
        columns=flexible
    }
]

You are analyzing a user's operation log on an Android device. \\
Your goal is to integrate a NEW Memory Event into an EXISTING Memory Event. You must merge them into a single, cohesive narrative that reflects the entire journey within
this app session.\\
1. Integration Rules (Seamless Narrative):\\
- **Identify Continuity**: If the new event continues the previous task (e.g., previously searched for shoes, now filtering by size), merge them into a single flow.\\
- **Capture Pivots**: If the user abandoned the previous task for something new (e.g., stopped looking at shoes and started searching for electronics), reflect this
transition (e.g., 'After looking at shoes, the user shifted focus to...').\\
- **Deduplication**: If the new event repeats information from the existing event, do not list it twice; prioritize the most recent state or progress.\\
2. Details Merging Rule:\\
- Combine the 'Existing Details' with 'New Details' into one comprehensive list.\\
- **Conflict Resolution**: If a value has changed (e.g., quantity updated from 1 to 2), keep the most recent data point.\\
- Ensure all hard data (prices, codes, specs) from both sets are preserved unless they are outdated.\\
3. State Re-evaluation:\\
- **finished**: The combined journey has now reached a terminal point (e.g., payment completed, final result obtained).\\
- **ongoing**: The task remains incomplete, or the user is still in the middle of a new workflow.\\
- **Note**: Even if the previous state was 'ongoing', it can now become 'finished' if the new event completes the task.\\
Output Format:\\
Event: <A single integrated summary combining existing and new events>\\
Details: <Merged list of all specific data points (Numeric values, Proper nouns, Specific attributes).>\\
State: <finished or ongoing>\\
Example: \{Example\}\\
Input Data to Merge:\\
App: \{app\_name\}\\
**[Existing Context]**\\
Existing Event: \{existing\_event\}\\
Existing Details: \{existing\_details\}\\
**[New Information]**\\
New Event: \{new\_event\}\\
New Details: \{new\_details\}
\end{tcolorbox}

\begin{tcolorbox}[
    float*=t,                 
    width=\textwidth,         
    title=Prompt Template for Application Inference,
    breakable,
    listing options={
        basicstyle=\ttfamily\small, 
        breaklines=true,            
        columns=flexible
    }
]

You are an agent skilled at analyzing user goals and identifying missing information.
Given the user's goal: \{goal\}\\
\{app\_desc\}\\
\{missing\_info\}\\
Which apps might contain useful information for completing this goal? Only return apps from the list above.
"Return the app names in the following format:\\
app 1: <app\_name1>\\
app 2: <app\_name2>\\
...\\
You MUST return at least one app that you think might contain useful information.
\end{tcolorbox}

\begin{tcolorbox}[
    float*=t,                 
    width=\textwidth,         
    title=Prompt Template for Information Filtering,
    breakable,
    listing options={
        basicstyle=\ttfamily\small, 
        breaklines=true,            
        columns=flexible
    }
]

You are a precise context filter for an Android GUI Agent.
Given the user's goal:\\
\{goal\}\\
\{missing\_info\}\\
And the following potentially useful memory items (in JSON format):\\
\{memory\_json\}\\
Extract the information that is useful for completing this goal.\\
Return a concise description of the relevant information.
This subsequent information will serve as supplementary content to accomplish this
goal. If the history contains no information that directly aids the Goal, return an empty
string.
\end{tcolorbox}

\begin{tcolorbox}[
    float*=t,                 
    width=\textwidth,         
    title=Prompt Template for Sufficiency Verification,
    breakable,
    listing options={
        basicstyle=\ttfamily\small, 
        breaklines=true,            
        columns=flexible
    }
]

You are evaluating whether the retrieved information is sufficient for an Android
phone agent to complete a user's goal. The agent can interact with the phone's UI (tap
, swipe, type, search within apps, etc.).\\
**What the agent can do on its own (does NOT need from memory):**\\
- Open apps by name\\
- Search for contacts or items by name within apps\\
- Navigate UI elements and select items by visible text\\
- Perform standard app operations\\
**What the agent DOES need from memory (information it cannot directly discover
through UI):**\\
- Specific data values that must be recalled exactly (codes, passwords, amounts, etc.)\\
- Content the user previously saw or entered that is no longer on screen\\
- Cross-app facts that the agent cannot infer without prior context\\
Given the user's goal: {goal}\\
And the following retrieved information:\\
\{retrieved\_info\}\\
Determine if the retrieved information provides the data the agent needs, or if critical information is still missing. Be conservative: if the goal only requires actions the agent can perform through UI interaction, and no specific data values are needed from memory, mark as sufficient. Return your answer in the following format:\\
sufficient: yes/no\\
missing information: <describe what specific data is still missing, if any. Only list information the agent truly cannot obtain through UI interaction.>

\end{tcolorbox}
\clearpage
\bibliography{aaai2027}


\end{document}